\documentclass[letterpaper, 10 pt, conference]{ieeeconf}  % Comment this line out if you need a4paper

\IEEEoverridecommandlockouts                              % This command is only needed if 
\makeatletter
    \let\NAT@parse\undefined
\makeatother
\usepackage[square, numbers, sort&compress]{natbib}
\usepackage{graphicx}
\usepackage{colortbl}
\usepackage{booktabs}
\usepackage{comment}
\usepackage{times}
\usepackage{epsfig}
\usepackage{lipsum}
\usepackage{amsmath}
\usepackage{pifont}
\usepackage{amssymb}
\usepackage{arydshln}
\usepackage{color}
\usepackage{caption}
\usepackage{url}
\usepackage{here}
\usepackage{balance}
\usepackage{xspace}
\usepackage{multirow}
\usepackage{bbding}
\usepackage[ruled,vlined]{algorithm2e}
\usepackage[bookmarks=true, pagebackref]{hyperref}
\let\labelindent\relax
\usepackage{enumitem}

\newcommand{\colornewparts}[0]{\color{black}}

\newcommand{\cmark}{\ding{51}}%
\newcommand{\xmark}{\ding{55}}%

\newcommand{\MethodName}[0]{SACSoN\xspace}
\newcommand{\DataName}[0]{HuRoN\xspace}

\renewcommand{\baselinestretch}{1.0}

\title{\LARGE \bf
\MethodName: Scalable Autonomous Control for Social Navigation}%Scalable Autonomous Data Collection for Social Navigation

\author{Noriaki Hirose$^{1,2}$, Dhruv Shah$^{1}$, Ajay Sridhar$^{1}$ and Sergey Levine$^{1}$
\thanks{$^{1}$UC Berkeley, $^{2}$Toyota Motor North America}%
}
\def\eqref#1{equation~\ref{#1}}
\def\1{\bm{1}}

\DeclareMathAlphabet{\mathsfit}{\encodingdefault}{\sfdefault}{m}{sl}
\SetMathAlphabet{\mathsfit}{bold}{\encodingdefault}{\sfdefault}{bx}{n}

\def\gH{{\mathcal{H}}}

\def\gR{{\mathcal{R}}}

\begin{document}
\maketitle
\thispagestyle{empty}
\pagestyle{empty}

%%%%%%%%%%%%%%%%%%%%%%%%%%%%%%%%%%%%%%%%%%%%%%%%%%%%%%%%%%%%%%%%%%%%%%%%%%%%%%%%
\begin{abstract}

Machine learning provides a powerful tool for building socially compliant robotic systems that go beyond simple predictive models of human behavior. By observing and understanding human interactions from past experiences, learning can enable effective social navigation behaviors directly from data. 
{\colornewparts In this paper, our goal is to develop methods for training policies for socially unobtrusive behavior, such that robots can navigate among humans in ways that don't disturb human behavior in visual navigation using only onboard RGB observations.} We introduce a definition for such behavior based on the \emph{counterfactual} perturbation of the human: if the robot had not intruded into the space, would the human have acted in the same way? By minimizing this counterfactual perturbation, we can induce robots to behave in ways that do not alter the natural behavior of humans in the shared space. Instantiating this principle requires training policies to minimize their effect on human behavior, and this in turn requires data that allows us to model the behavior of humans in the presence of robots. Therefore, our approach is based on two key contributions. First, we collect a large dataset where an indoor mobile robot interacts with human bystanders. Second, we utilize this dataset to train policies that minimize counterfactual perturbation.
We provide supplementary videos and make publicly available the visual navigation dataset on our project page\footnote{\bf \href{http://sites.google.com/view/SACSoN-review}{\texttt{
sites.google.com/view/SACSoN-review}}}.

\end{abstract}

%%%%%%%%%%%%%%%%%%%%%%%%%%%%%%%%%%%%%%%%%%%%%%%%%%%%%%%%%%%%%%%%%%%%%%%%%%%%%%%%
\section{Introduction}

Even the simplest forms of interaction between humans, such as how to pass someone in a hallway, are governed by complex non-verbal cues, and may be challenging to script. In order for robots to inhabit the same environments as people, they must also be cognizant of basic social cues and etiquette, even for seemingly simple navigational tasks. While a range of prior works have proposed approaches for modeling human behavior~\cite{helbing1995social,ferrer2013robot}, the complexity of such interactions often defies analytic modeling techniques.

%{\colornewparts {\bf TODO modify this paragraph, because I almost copy and paste the sentences from abstract.}}
We approach this challenge from a data-driven perspective: acquiring policies for navigation around humans by leveraging data of human-robot interactions to \emph{learn} how to navigate in socially unobtrusive ways. We propose a definition for such behavior, which is based on the \emph{counterfactual} perturbation of humans. Specifically, we consider whether humans would have acted in the same way if the robot had not intruded into their space. By minimizing this counterfactual perturbation, we can guide robots to behave in a manner that does not alter the natural behavior of humans in the shared space. {\colornewparts To instantiate this principle, we train the \MethodName ({\bf S}calable {\bf A}utonomous {\bf C}ontrol for {\bf So}cial {\bf N}avigation) policy to minimize the impact on human behavior for vision-based navigation using a single camera.} This requires us to both formalize the notion of counterfactual perturbation into an objective, and to collect a dataset that has the kinds of human-robot interactions that can allow our model to learn to predict human behavior in the presence of robots. Thus, our work focuses on two complementary technical components: the design of a policy learning method that can utilize predictive models of humans for unobtrusive navigation, and the collection of a large dataset of human-robot interactions to train these predictive models. % Hence, our approach is built upon two essential contributions. Firstly, we collect a substantial dataset in which an indoor mobile robot interacts with human bystanders. Secondly, we leverage this dataset to train policies that effectively minimize counterfactual perturbation.
%
%, but can instead learn how to either forecast human behavior or even acquire socially cognizant navigational strategies from data. 
%{\colornewparts The most effective modern learning techniques typically require large and representative datasets, but in robotics collecting real-world data tends to use manual teleoperation, which is expensive and scales poorly. Besids, autonomously collected dataset generally does not contain the robotic command including human experience and knowledge to learn and will be less diverse.
%Indeed, human-interactive systems have been successfully learned from data and experience in a range of other domains, such as playing games~\cite{silver2017mastering,vinyals2017starcraft,kramar2022negotiation} and generating dialogue~\cite{mirowski2022co,stiennon2020learning}. However,  and oftentimes the larger the dataset the more effective the resulting system can be. 
%Therefore, in this paper we ask the question: \emph{How can we scalably collect large datasets of human-robot interaction and learn the socially compliant control policy from the autonomously collected dataset?}}

\begin{figure}[t]
  \begin{center}
      \includegraphics[width=0.9\hsize]{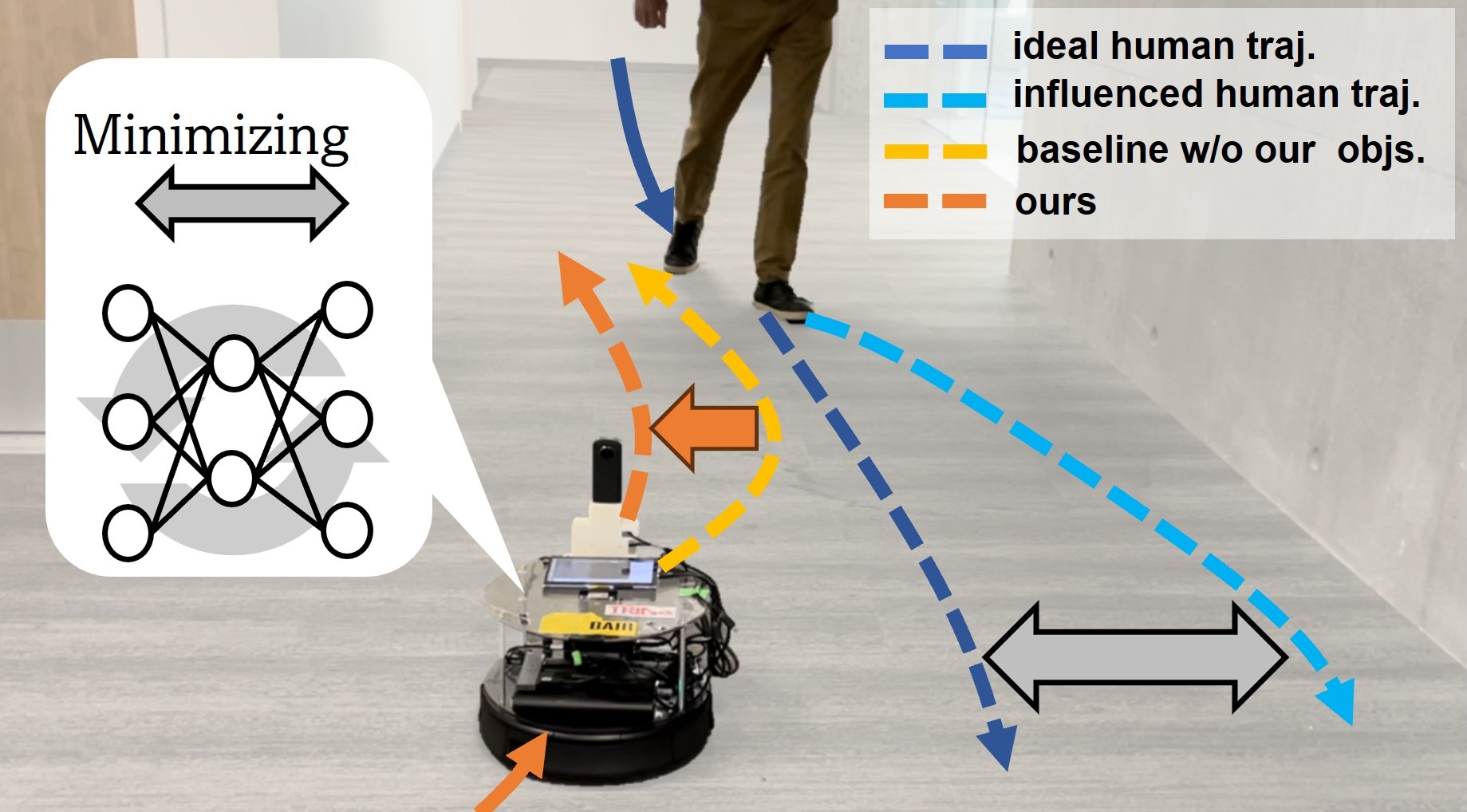}
  \end{center}
	\caption{\small {\bf \MethodName is a socially unobtrusive vision-based navigation policy in the human-occupied spaces.} We penalize counterfactual perturbations (gray) from the intended human trajectory (navy) and generate the compliant commands (orange).}
  \label{f:pull}
  \vspace*{-1.5em}
\end{figure}

To collect such a dataset, we propose a data collection system, which we call \DataName ({\bf Hu}man-{\bf Ro}bot interaction data collection for vision-based {\bf N}avigation) system.
%%SL.7.13: define what the acronym means?
%%NH.7.15: I did in the previous paragraph.
In contrast to previous social navigation datasets that involve expensive manual tele-operation~\cite{martin2021jrdb, karnan2022socially}, or simple scripted policies that fail to capture data diversity~\cite{thorDataset2019}. Instead, we devise an intelligent system that can \emph{autonomously} collect rich interaction data with little-to-no human intervention, and can improve its data collection policy over time as the ever-growing dataset is reused to further train our policy.
\begin{table*}[t]
  \begin{center}
  \vspace{0mm}
  \caption{{\small {\bf Survey of public datasets for learning vision-based navigation policies in real-world environments around humans.} \DataName is the largest available visual navigation dataset of an autonomous policy interacting with humans in real-world environments.}}  
  \label{tab:data}    
  \resizebox{2.0\columnwidth}{!}{
  \begin{tabular}{lccccl} \toprule 
    Dataset & Human & Policy & Duration [hour] & Distance [km] & Sensors\\ \midrule
    KITTI odometry~\cite{Geiger2012CVPR} & \xmark & teleop & 0.7 & 22.2 & stereo RGB, 3D LiDAR, GPS \\
    NCLT~\cite{carlevaris2016university} & \xmark & teleop & 34.9 & 147.4 & RGB, 3D LiDAR, odom, GPS, IMU\\ 
    GO Stanford~\cite{hirose2019deep} & \xmark & teleop & 10.3 & 16.7 & spherical RGB, odom\\
    FLOBOT~\cite{yan2020robot} & \xmark & autonomous & 0.46 & 0.2 & RGBD, stereo RGB, 3D and 2D LiDAR, odom, IMU.\\
    RECON~\cite{shah2021rapid} & \xmark & autonomous & 25.0 & 81.0 & stereo fisheye RGBD, thermal, 2D LiDAR, GPS, IMU\\
    JRDB~\cite{martin2021jrdb} & \cmark & teleop & 1.1 & 2.3 & stereo RGBD, 3D and 2D LiDAR, IMU\\
    SCAND~\cite{karnan2022socially} & \cmark & teleop & 8.7 & 40.0 & RGBD, 3D LiDAR, odom\\ 
    TH{\"O}R~\cite{thorDataset2019} & \cmark & scripted & 1.0 & 1.0 & 3D LiDAR, motion capture, eye-tracking glasses\\
    MuSoHu~\cite{nguyen2023toward} & \cmark & no robots & {\colornewparts 20.0} & {\colornewparts 100.0} & spherical RGB, stereo RGBD, 3D LiDAR, IMU\\ \midrule    
    \DataName (Ours) & \cmark & autonomous & 75.0 & 58.7 & spherical RGBD, fisheye RGB, 2D LiDAR, odom, bumper\\\bottomrule
  \end{tabular}%
  }
  \end{center}
  \vspace*{-1.5em}
\end{table*}

We deploy our data collection system to collect the \DataName dataset, which comprises over 75 hours of robotic navigation in 5 different office environments populated by people. To the best of our knowledge, this represents the largest
%%SL.2.28: the largest public one? (well, if it's actually public...); what about cobot?
%%NH.3.1: cobot does not include RGB images.
such dataset of an autonomous mobile robot interacting with humans, with over 4000 individual human-robot interactions. In the process of collecting the \DataName dataset, our robot traveled for a combined total of about 58.7 km over four months. Since our dataset includes time sequences of camera images, 2D LiDAR, and wheel odometry, our dataset can be useful for visual SLAM tasks including visual odometry estimation and depth estimation.

{\colornewparts Our work makes the following contributions: (i) a model-based method for learning a socially compliant \MethodName policy for visual navigation around humans%(Note that our model-based method does not belong to reinforcement learning.)
}, (ii) an autonomous data collection system, \DataName, that encourages rich interactions with human pedestrians using a novel training objective, and (iii) the \DataName dataset, a large and diverse dataset comprising over 4000 human-robot interactions of an autonomous robot operating in a densely populated office-space environment. Please see the project page for the dataset and videos.
%The primary contributions of this paper are three-fold: (i) an autonomous data collection system, the \MethodName system, that encourages rich interactions with human pedestrians using a novel objective, (ii) the \MethodName dataset, a large and diverse dataset comprising over 4000 human-robot interactions of an autonomous robot operating in a densely populated office-space environment, {\colornewparts and (iii) novel objectives to learn the \MethodName policy for vision-based navigation that allow robots to navigate among humans with less disruption.} Please see the project page for the dataset release, additional implementation details and videos.
%{\colornewparts We show that our interaction-enriched dataset can be used for a variety of downstream tasks, including socially compliant navigation policies for mobile robots and human motion forecasting.}
%{\colornewparts To the best of our knowledge, this is the largest publicly available dataset of visual navigation with social interactions publicly. And our proposed training approach enables us to leverage this largest dataset and allows to learn socially compliant policy in vision-based navigation. Please see the project page for the dataset release, additional implementation details and videos.}
%
\section{Related Work}

Social navigation has been widely studied in the literature~\cite{mavrogiannis2021core,sisbot2007human,mumm2011human}. Model-based approaches based on the dynamic pedestrian model have clasically been applied for behavior modeling~\cite{helbing1995social,ferrer2013robot,mehta2016autonomous}. These methods determine the robot's actions in a virtual space with the predicted pedestrians' behavior~\cite{luber2012socially,ziebart2009planning,bajcsy2019scalable,pfeiffer2016predicting,mavrogiannis2018social,xiao2022learning,truong2017approach,chen2023social,bhaskara2023sg,narayanan2023ewarenet,rosmann2017online,chen2018robot}, considering social momentum~\cite{mavrogiannis2018social}, a maximum entropy model~\cite{pfeiffer2016predicting}, a model predictive controller~\cite{xiao2022learning}, or a classical planner~\cite{truong2017approach,chen2023social}.
%, and more recently in the context of learning forecasting models of pedestrian behavior~\cite{alahi2016social,sadeghian2019sophie}. 
Social navigation has also been viewed through the lens of model-free data-driven learning such as reinforcement learning~\cite{chen2017socially,chen2017decentralized,everett2018motion,chen2019crowd,mun2022occlusion}. 

Our method using the pedestrians' predictive model belongs to the former. {\colornewparts However, different from prior works, including model-based reinforcement learning, we apply the predictive model to estimate the counterfactual perturbation from the pedestrians' intended trajectory and train the control policy offline by penalizing the perturbations.} Hence our control policy enables the robot to navigate to the target position while allowing the pedestrian to walk as intended. {\colornewparts Moreover, since our approach is end-to-end learning, the robot actions can be derived from raw images without detecting and predicting pedestrians in inference.}
%However, such approaches tend to rely on high-fidelity simulated environments and simulated pedestrian dynamics models. Instead, we focus on learning vision-based navigation policies for social interactions via the dynamic pedestrian model from real-world interactions~\cite{pokle2019deep,luber2012socially,karnan2022socially}.
%
%\cite{luber2012socially,ziebart2009planning} uses the predictive model to determine the robot actions. In the predicted virtual space, they determine the commands with the estimated pedestrians behavior. 
%\cite{bajcsy2019scalable} is for drone. They determines the commands with the predicted human motions. 
%\cite{pfeiffer2016predicting} builds a system devloped in a maximum entropy distribution using the predictive model.
%\cite{mavrogiannis2018social} designs the social momentum planning framework for socially-aware multi-agent navigation.
%\cite{poddar2023crowd,xiao2022learning} uses mpc aproach.
%\cite{truong2017approach,chen2023social} uses the predictive model and determine the robot actions.

Similar to the data-driven approaches, our training method {\colornewparts needs a large dataset}.
%to have accurate the predictive model for the pedestrians' behavior as well as our control policy. 
%For vision-based navigation, we often use photorealistic simulators~\cite{savva2019habitat,xia2018gibson,savva2017minos,kolve2017ai2} or public dataset collected in the real environments~\cite{hirose2019deep, shah2021ving, karnan2022socially, levine2023learning}.
%supervised data collection~\cite{francis2020long} to learn goal-reaching policies directly from raw visual observations. Instead, we focus on learning visual navigation policies from autonomously collected real-world experience, which can in principle improve continuously as more data is collected~\cite{hirose2019deep, shah2021ving, karnan2022socially, levine2023learning}. 
For vision-based navigation, prior works in collecting real-world data tend to use manual teleoperation, which is expensive and scales poorly~\cite{Geiger2012CVPR, carlevaris2016university, hirose2019deep, karnan2022socially, martin2021jrdb}. Instead, in addition to the training method, our work focuses on autonomous data collection of rich human-robot interactions, aiming to train our control policy.
{\colornewparts While there has been prior work on autonomously collecting robot navigation data~\cite{thorDataset2019, biswas2013localization, yan2020robot, shah2021rapid},
our task is particularly challenging due to the dynamic agents (i.e., humans) present in the environment.
To autonomously learn an accurate predictive model and socially-compliant behavior around humans, the training data must contain rich human-robot interactions, with humans walking close to the robot, and it must include a wide perceptual and behavioral diversity. The closest prior works are SCAND~\cite{karnan2022socially}, which is teleoperated in indoor and outdoor environments, and CoBoT, TH\"OR~\cite{biswas2013localization, thorDataset2019}, which are autonomous but contain no visual observations; therefore, they have limited utility for learning visual navigation policies. MuSoHu~\cite{nguyen2023toward} collects a dataset without using real robots. Instead, they have human participants walk in human-occupied spaces. Hence, they do not include any interactions between real robots and humans.
%% TODO: Can you add the references that I added to rebuttal doc regarding this point?
Table~\ref{tab:data} summarizes the existing robot navigation datasets, highlighting scene, method, size, and contained sensor signals. In addition to the training method, we propose the \DataName system that can autonomously collect a large and diverse dataset of rich human interactions, and can be scaled with minimal human effort to multiple environments.}

%%SL.7.14: The current related work section actually has no discussion of prior work on objectives -- is there no prior work at all that proposes any objective similar to our counterfactual objective? It feels like there must be *something* that is at least somewhat related, and adding a few sentences about prior works that have related objectives would really help.
%%NH.7.17: I added some prior works using the pedestrian's prediction and highlighted the dffierence from them in first and sencond paragraphs.
%
%
\section{Preliminaries}

%%SL.7.14: This section starts off as though the paper is only about the data collection system, can we update this to account for the new paper story we have?
%%NH.7.15: I edited the first paragraph.
We propose a method and dataset for social compliant robotic navigation with a learning-based approach.
The design of our method extends ExAug~\cite{hirose2022exaug}, a control policy for vision-based navigation that optimizes a goal-directed cost function (but does not by itself consider interaction with humans). This system can navigate to user-specified goal images using a combination of a topological graph and a learned low-level control policy, and its design is related to a number of recent works on vision-based navigation with learned policies and topological maps~\cite{savinov2018semi,meng2020scaling,shah2022viking,hirose2019deep,hirose2022exaug,kim2022topological}.
We build our data collection system, \DataName, on top of the same visual navigation system.
%We develop our data collection system (Fig.~\ref{f:block}) on top of a visual navigation system that can navigate to user-specified goal images using a combination of a topological graph and a learned low-level control policy. 
%This method has been demonstrated in a variety of prior works~\cite{savinov2018semi,meng2020scaling,shah2022viking,hirose2019deep,hirose2022exaug,kim2022topological}. Specifically, we build our data collection system on top of ExAug~\cite{hirose2022exaug},  that uses a model-based control policy at the low level to optimize goal-directed cost function.

The control policy in ExAug predicts control velocities $\{v_i, \omega_i \}_{i=1 \ldots N_s} = \pi^{c}_{\phi}(I_t, I_g)$
%%SL.7.14: Do we need the subscript "_c" for "theta_c"? Can we just call it theta to keep the notation simpler?
%%NH.7.14: OK. I put the other methematical symbols.
from the current image $I_t$ and subgoal image $I_g$, and commands the linear velocity $v_1$ and the angular velocity $\omega_1$ to the robot to reach the position of $I_g$, similar to receding horizon control. Here, $N_s$ is the control horizon and $t$ is the current step number. We commonly show the learnable parameters (e.g., $\phi$) as a subscript on the model function (e.g., $\pi^{c}_{\phi}$). The control policy is paired with a topological memory that contains images as nodes and temporal distance between them as the edges. The ExAug control policy $\pi^{c}_{\phi}$ is trained 
%%SL.2.28: I think we should try to explain this a little more. Many readers will probably interpret "supervised learning" to mean "imitation learning" in this context. It's not really supervised learning though, it's just directly optimizing the objective below. There is a bit of setup that is needed to understand this in terms of how ExAug actually works (i.e., it produces a trajectory and forms these objective terms as differentiable collision/pose/etc objectives such that the policy can be optimized directly with backprop)
%%SL.7.14: Would be good to address the above comment.
%%NH.7.15: I added one sentence after equation (1) to clarify that ExAug is not imitating the GT velocity commands.
to minimize the objective
\begin{equation}
\,J_\text{nav}(\phi) := J_\text{pose}(\phi) + w_c J_\text{col}(\phi) + w_r J_\text{reg}(\phi),
\end{equation}
where $J_\text{pose}$ corresponds to the prediction error in the relative pose estimates, $J_\text{col}$ penalizes collisions, and $J_\text{reg}$ is a regularization term for predicted velocities. 
{\colornewparts ExAug uses a geometric and kinematics model to estimate the relevant states of the robot in a virtual space and calculate these objectives, akin to the model predictive control. These objectives enable us to train the policy by minimizing the differentiable cost $J_\text{nav}$ without imitating the ground truth values.} Please refer to the original paper for implementation details of this system~\cite{hirose2022exaug}. 

\vspace{1mm}
\noindent
\textbf{Overview:} 
%{\bf TODO add small overview here about organization of the next two sections.}
Section \ref{sec:soc_nav} introduces our method to train the \MethodName policy, which aims to enable robotic navigation among humans with minimal disruption. In addition to $J_\text{nav}$, we introduce two new objectives using the counterfactual human trajectories. We pre-train the predictive model of the pedestrians' future trajectory to estimate the counterfactual human trajectories in training. 
%%SL.7.14: The notion of a predictive model of humans wasn't introduced yet, so the above sentence sounds a bit weird.
%%NH.7.15: I modified to explain the predictive model a bit.
Section \ref{sec:data_collection} describes the \DataName system for autonomously collecting a dataset with human-robot interactions that allows us to effectively train the \MethodName policy. Our data collection system includes two key components. First, we use a policy that is similar to \MethodName, but optimized to \emph{encourage} rather than \emph{avoid} interactions with humans, so as to gather the maximum number of human-robot interactions. Second, the \DataName system is designed for scalable, autonomous data collection, and includes a number of design choices to enable autonomy and continual improvement that we detail in Section \ref{sec:data_collection}.
%\DataName system is built upon ExAug along two key axes: by encouraging interactions with human pedestrians using a different objective only for \emph{data collection purposes}, and automating the deployment of the navigation system for scalable data collection without human interventions with three components: a help-and-rescue module, long-term anchors, and continual learning.%(Sec.~\ref{sec:data_collection}). To collect data rich in human-robot interaction, we propose a different objective to encourage the robot to interact with pedestrians only for \emph{ data collection purposes}. We also introduce three components to our scalable data collection system that minimize the number of human interventions: a help-and-rescue module, long-term anchors, and continual learning.

\begin{figure}[t]
  \begin{center}
      \includegraphics[width=0.99\hsize]{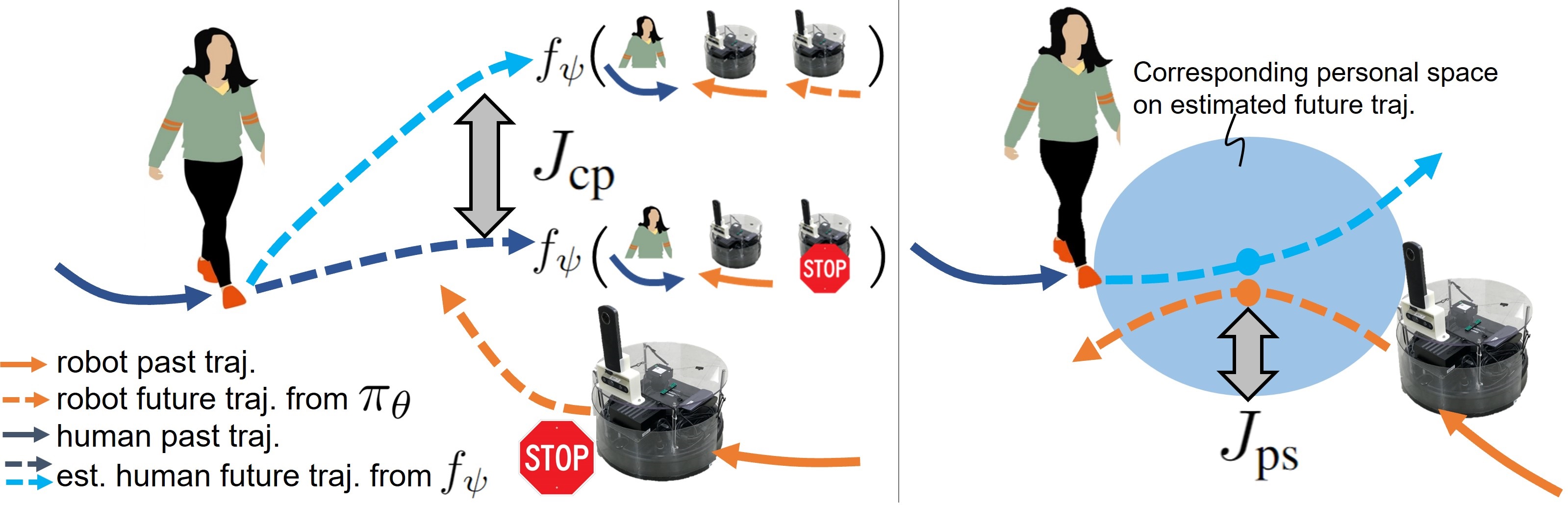}
  \end{center}
      \vspace*{0mm}
	\caption{\small {\bf Our proposed objectives $J_\text{cp}$ and $J_\text{ps}$ for training \MethodName policy.} $J_\text{cp}$ penalizes the counterfactual perturbation from the estimated intented pedestrian's trajectory (left). $J_\text{ps}$ penalizes the personal space violation in the future space (right).}
  \label{f:objective}
  \vspace*{-1.5em}
\end{figure}
\section{Learning a Socially Compliant Policy}
\label{sec:soc_nav}
%{\bf TODO modify intro and motivation following language from abstract.}
We posit that a possible way to achieve ``social compliance'' is for robots to avoid disrupting the \emph{intended behavior} of pedestrians, i.e., allow humans to carry on with their activities without disruption. 
In our proposed method, we penalize the counterfactual perturbation of the intended trajectories of the pedestrians. We define the intended trajectory of a pedestrian as the predicted trajectory of the pedestrian from our predictive model conditioned on the robot being stationary and non-intrusive. Our method aims to control the robot so that the humans in the environment do not act differently than they would have if the robot had been stationary. This principle could be further generalized to minimize the difference to other counterfactual situations, such as ones where the robot is absent all together, but we focus on the stationary robot counterfactual as a simple instantiation of the principle.
%%SL.7.14: Both here and in the overview we mention the predictive model -- can we introduce the predictive model in the preliminaries? That will make this a bit clearer.
%%NH.7.15: DONE.
%In our proposed method, we predict two counterfactual trajectories for the pedestrian: one conditioned on the trajectory of a stationary robot and one conditioned on the trajectory of a robot that takes our policy's proposed actions. We penalize the perturbation of the pedestrian's trajectory using these two counterfactual pedestrian trajectories in training to obtain a social compliant control policy. 
For safety, the complete design of our full objective function also includes a term to penalize the predicted distance between the human and the robot, to encourage the robot to maintain clearance, as well as the standard navigation terms described in the preceding section.
%One of the simplest forms of socially compliant behavior is to avoid intruding someone's personal space and give way to the pedestrians. Since the robot is generally slower than the pedestrians in indoor, giving way is simple but plausible behavior. In this section, we show our approach learning social etiquette of this form from \MethodName dataset. 
Thus, we add two terms to $J_\text{nav}$, forming our full objective:
\begin{equation}
    \min_{\theta} \,J(\theta) := J_\text{nav}(\theta) + w_{cp}J_\text{cp}(\theta) + w_{ps}J_\text{ps}(\theta),
    \label{eq:objective2}
\end{equation}
%\begin{equation}
%    \min_{\theta} \,J(\theta) := J_\text{nav}(\theta) + J_\text{cp}(\theta) + J_\text{ps}(\theta),
%    \label{eq:objective2}
%\end{equation}
%%SL.7.14: why are there two objectives differing only in the presence/absence of weights? is that a mistake, should there be just one?
%%NH.7.14: Just mistake! Sorry!
%
where $J_\text{cp}$ is an objective to suppress the counterfactual perturbation (Fig.~\ref{f:objective} left) and $J_\text{ps}$ is an objective to penalize the penetration of the personal space of the pedestrians (Fig.~\ref{f:objective} right), where $w_{cp}$ and $w_{ps}$ are weights for each objective.
%, and $\theta$ represents the learnable parameters of our control policy. 
Here, our control policy $\pi_\theta$ predicts velocity commands $\{v_i, \omega_i\}$ from $I_{t:t-N_p}$ and $I_g$, defined as follows:
%%SL.7.14: should this be in the preliminaries section?
%%NH.7.14: The preliminary explains ExAug, which does not need the past images. But, I tried to put these explanation in the preliminary.
%
\begin{equation}
    \{v_i, \omega_i \}_{i=1 \ldots N_s} = \pi_\theta(I_{t:t-N_p}, I_g)
    \label{eq:control_policy}
\end{equation}
Concatenating the past image frames gives the robot additional context that can be useful to avoid obstacles, detect pedestrians in the environment, and reduce partial observability~\cite{shah2022gnm}.
\begin{figure}[t]
  \begin{center}
      \includegraphics[width=0.99\hsize]{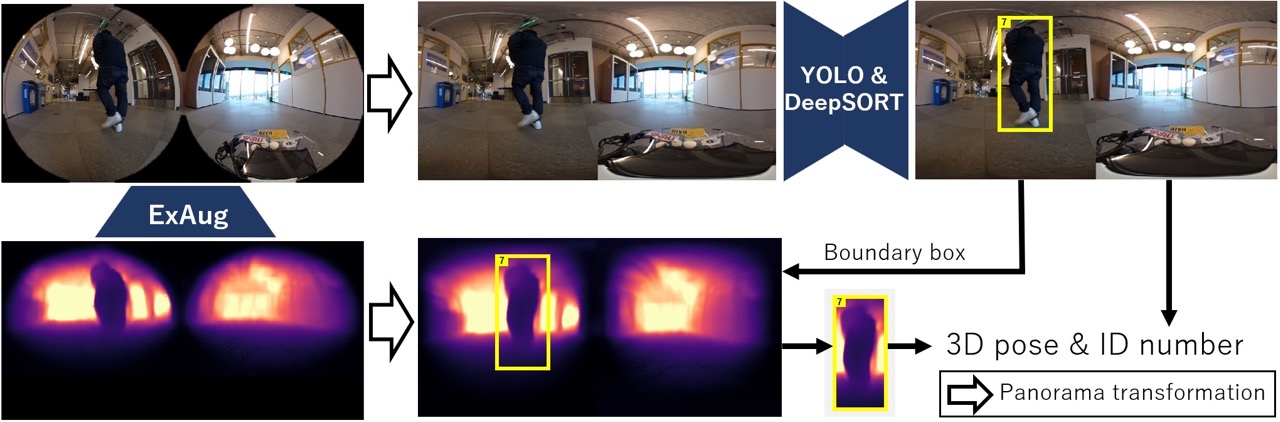}
  \end{center}
      \vspace*{0mm}
	\caption{\small {\bf Pedestrian detection and tracking.} We use a combination of YOLO and DeepSORT to detect and track pedestrians from visual observations, and estimate their relative position using the scaled depth estimates from ExAug's perception module.}
  \label{f:pose_est}
  \vspace*{-1.5em}
\end{figure}

\noindent \textbf{$J_\text{cp}$:}
%The easiest way to give way to pedestrians is stopping when detecting the pedestrians around the robot. However, its behavior can not allow the robot to go to the goal position, if the pedestrians stop and keep to be appeared around the robot. To handle both giving way and reaching goal, 
To train the policies without distracting pedestrians, we design $J_\text{cp}$ using counterfactual pedestrians trajectories,
\begin{equation}
    J_\text{cp}(\theta) = \frac{1}{N_s}\sum_{i = 1}^{N_s}(\hat{h}^{gw}_{t+i} - \hat{h}_{t+i})^2,
    \label{eq:objective3}
\end{equation}
where $\hat{h}^{gw}_{t+i}$ is the estimated pedestrian's 2D trajectory conditioned on the robot virtually stopping at the current position to give way and $\hat{h}_{t+i}$ is the estimated pedestrian's 2D trajectory conditioned on the robot future action. By minimizing $J_\text{cp}$ with the other objectives to train our control policy, the pedestrian can walk a path similar to what they would have taken when the robot stopped and gave way, while allowing the robot itself to move toward the goal position. Here, we estimate $\hat{h}_{t+i}$ as
\begin{equation}
\begin{split}
    %\hat{h}^{gw}_{t:t+\beta} &= f_{\psi}(h_{t-\alpha:t-1}, r_{t-\alpha:t-1}, {\bf 0}) \\
    \hat{h}_{t+1:t+\beta} &= f_{\psi}(h_{t-\alpha:t}, r_{t-\alpha:t}, r_{t+1:t+\beta})
\end{split}    
    \label{eq:prediction}
\end{equation}
where $f_{\psi}$ is a trained predictive model of a pedestrian's future trajectory, conditioned on their past trajectory $h_{t-\alpha:t}$, as well as the robot's past trajectory $r_{t-\alpha:t}$ and future trajectory $r_{t+1:t+\beta}$. All trajectories in Eqn.~\ref{eq:prediction} are on the current robot coordinate. The values for $r_{t-\alpha:t}$ are obtained from past wheel odometry, and $r_{t+1:t+\beta}$ is derived by integrating the velocity commands $\{v_i, \omega_i \}_{i=1 \ldots N_s}$ from our control policies. To obtain $h_{t-\alpha:t}$, we use YOLO~\cite{redmon2016you,yolo_v5} and DeepSORT~\cite{wojke2017simple} to detect and track pedestrians in the images (processed into a panorama) from the recorded observations of the robot~\cite{yolo_deepsort}, and project these detections in 3D using the depth and scale estimates~\cite{niwa2022spatio} obtained from the ExAug perception module, as shown in Fig.~\ref{f:pose_est}.

For the other counterfactual trajectory, we input a zero vector instead of $r_{t+1:t+\beta}$
%%SL.7.14: is r a position or velocity? we said before it's formed by integrating velocity commands, so presumably it's a position, and is not zero?
%%NH.7.15: r is position. And the corrdinate of r is on the current robot coordiate. So, zero corresponds to stop at the current position. I mentioned after eq.(5).
to estimate $\hat{h}^{gw}_{t+1:t+\beta}$ as $f_{\psi}(h_{t-\alpha:t}, r_{t-\alpha:t}, {\bf 0})$. Giving the zeros vector as the robot future trajectory corresponds to stopping at the current pose. Note that we only consider scenes involving a single pedestrian for simplicity; for scenes with multiple pedestrians, we consider the nearest non-stationary pedestrians for training, since they are most likely to interact with the robot. To obtain an accurate predictive model $f_{\psi}$, we collect an interaction-enriched dataset using the \DataName system (Section \ref{sec:data_collection}), and train $f_{\psi}$ before training $\pi_\theta$.

%%SL.7.14: Can we add a couple of sentences to the above discussion to make it clear that this f function is the thing that is trained from data, hence the data collection is needed precisely to train f? Otherwise it's not clear where f comes from
%%NH.7.15: DONE
\noindent \textbf{$J_\text{ps}$:}
We design $J_\text{ps}$ to encourage the robot to avoid the personal space of the pedestrians.
\begin{equation}
    J_\text{ps}(\theta) = \mbox{min}_i \{\left|{\colornewparts \mathcal{R}_h +  \mathcal{R}_r} - c(d_i)\right|\},
    %J_\text{ps}(\theta) = \frac{1}{N_s}\sum_{i = 1}^{N_s}(r_h +  r_r - clamp(d_i))^2,
    \label{eq:objective4}
\end{equation}
where {\colornewparts $\mathcal{R}_h$} is the personal space, {\colornewparts $\mathcal{R}_r$} is the robot radius, $d_i$ is the distance on 2D plane between the future pedestrians' position $\hat{h}_{t + i}$ and the future robot position $r_{t+i}$, and $c$ is the function to limit $d_i$ between 0 and {\colornewparts $\mathcal{R}_h +  \mathcal{R}_r$} to penalize the robot trajectories only penetrating the personal space. $J_\text{ps}$ may be alternatively defined as the mean of the set $\{\left|{\colornewparts \mathcal{R}_h + \mathcal{R}_r} - c(d_i)\right|\}$, but empirically, we found the min formulation of Eqn.~\ref{eq:objective4} to better capture the desired behavior.
%
%Before training the control policy, we train $f_{\psi}$ by minimizing the MSE loss using supervised learning and frozen $f_{\psi}$ while training the control policy. We calculate the gradient of the control policy by back-propagation via $f_{\psi}$ for updating the control policy. For training, we design $w_{cp}$ and $w_{ps}$ as 10.0 and 100.0. All other hyperparameters are replicated from training process of data collection policy $\pi(\theta)$ in Section~\ref{sec:robot}.}
%
\begin{figure}[t]
  \begin{center}
      \includegraphics[width=0.99\hsize]{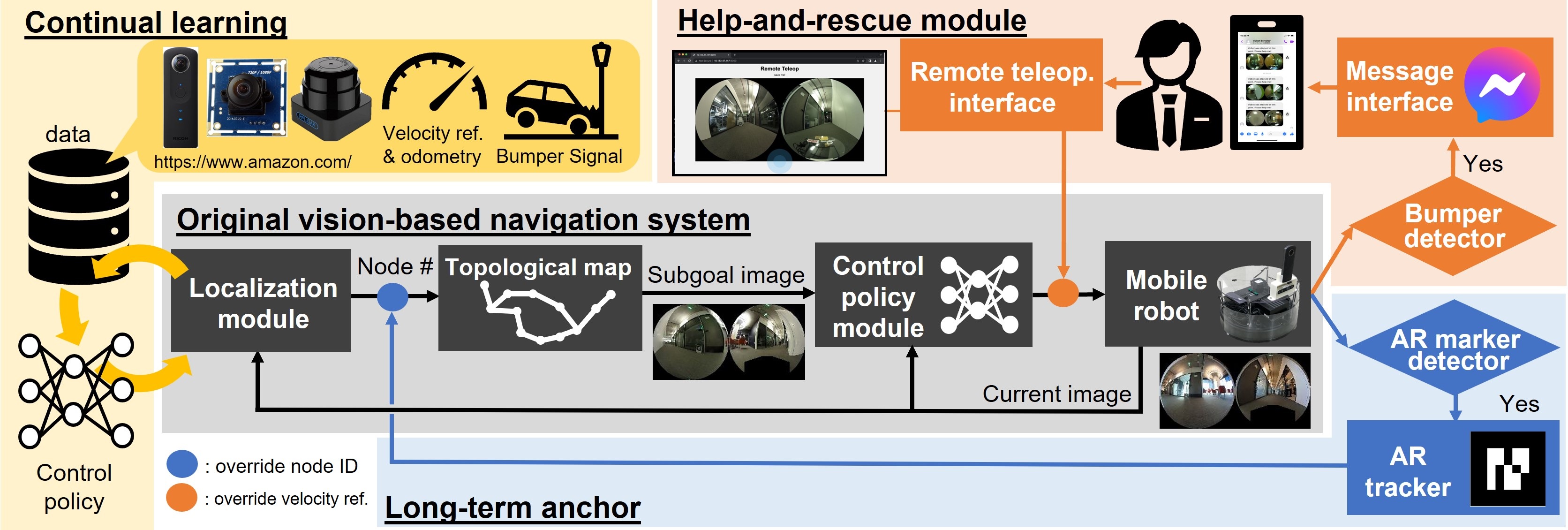}
  \end{center}
      \vspace*{0mm}
	\caption{\small {\bf \DataName System overview.} We design our autonomous data collection platform around a vision-based navigation system (gray) that uses a topological graph and a learned control policy. Our proposed system has three key components: a help-and-rescue module for collision recovery (orange), long-term anchors for localization (blue), and continual learning (yellow).}
  \label{f:block}
  \vspace*{-1.5em}
\end{figure}
\section{Autonomous Data Collection System}
\label{sec:data_collection}
%%SL.7.14: In general, this section is probably a bit too long now, let's try to shorten some of the details to get it more under control.
%%NH.7.15: I tried to shorten this section and move some parts into the appendix.
For our counterfactual objective to effectively supervise the robot's policy, we rely on the predictive model $f_{\psi}$ to make accurate predictions about hypothetical human-robot interactions. This requires training $f_{\psi}$ on a diverse dataset that contains many interactions between pedestrians and our robot. Therefore, the second major contribution of our work is an autonomous data collection system that can collect such a dataset. During collection, we wish to \emph{maximize} interactions between the robot and pedestrians, while also maintaining autonomy, to collect high-quality data.
%%SL.7.14: We should avoid starting a section with a subsection. What would really help is to put a paragraph here that motivates why we're making such a big deal out of the data collection, e.g. say something like: In order for our counterfactual objective to effectively supervise the robot's policy, we rely on the predictive model $whatever$ to make accurate predictions about hypothetical human-robot interactions. This requires training $whatever$ on a large and diverse dataset that contains many interactions between pedestrians and our robot. Therefore, the second major contribution of our work is an autonomous data collection procedure that we can use to collect such a dataset. During collection, we specifically aim to \emph{maximize} interactions between the robot and pedestrians, while maintaining maximal autonomy to collect the largest amount of data possible.
%%NH.7.15: Thank you!!
\subsection{System design}
We design a scalable data collection system with the data collection policy $\pi_{\rho}$ that is largely autonomous and can operate in large, indoor environments without any high-fidelity indoor positioning system.
Our proposed system (see Fig.~\ref{f:block}) builds on top of the existing ExAug navigation system using three key components: (a) help-and-rescue module for collision recovery, (b) long-term anchors for coarse localization in the environment, and (c) continual learning for improving performance over the course of deployment. 

\vspace{1mm}
\noindent
\textbf{Encouraging interactions:}
\label{sec:interactions}
%%SL.2.28: Maybe we should start with a broader overview, eg say that our goal is to collect data that [has something], and to that end we [do something], and then go into the details after that? That way it's clear to the reader what the goal of this section is
%
% To effectively train SACSoN policy in Section~\ref{sec:soc_nav}, SACSoN system in Fig.~\ref{f:block} collects the diverse dataset with enriched human-robot interactions to train more accurate predictive model $f_p$ as well as SACSoN policy $\pi_\theta$. We modify our visual navigation backbone to encourage human-directed interactions by proposing a novel objective that encourages the robot to drive closer to humans in the environment, causing humans to interact with the robot. We formulate this objective $J_\text{int}$ as the \emph{interaction loss}, that expresses the distance of the robot to the nearest pedestrian in the scene, and encourage the robot to minimize this objective during the training of our data collection policy. The modified training objective thus becomes
%
%{\bf TODO: add new text here}
In contrast to the \MethodName policy, the data collection policy $\pi_{\rho}$ is trained to collect a dataset with enriched human-robot interactions. 
%This dataset is then used to train the predictive model $f_\psi$ and the \MethodName policy $\pi_\theta$. 
We introduce an additional objective $J_\text{int}$ to encourage the robot to approach pedestrians while collecting data towards the desired goal.

\begin{equation}
    \min_{\rho} \,J(\rho) := J_\text{nav}(\rho) + w_i J_\text{int}(\rho),
    \label{eq:objective}
\end{equation}
where $w_i$ is a scaling factor. Here, we employs the same network structure $\pi_{\rho}$ as $\pi_\theta$ of Eqn.~\ref{eq:control_policy} for the data collection control policy. 
%
%predicts velocity commands $\{v_i, \omega_i\}$ and operates on the robot's past observations $I_{t:t-N_p}$, and the goal observation $I_g$, defined as follows:
%
%\begin{equation}
%    \{v_i, \omega_i \}_{i=1 \ldots N_s} = \pi_\theta(I_{t:t-N_p}, I_g)
%    \label{eq:control_policy}
%\end{equation}
%
%Concatenating the past image frames gives the robot additional context that can be useful to avoid obstacles, detect pedestrians in the environment and reduce partial observability~\cite{shah2022gnm}. % {\colornewparts Here, $\pi_\theta$ has same structure as $\pi^{c}_{\theta_c}$ except the number of the input channel.}
%
%{\colornewparts Mention here that the estimated depth HAS SCALE.}
%During training, we use YOLO~\cite{redmon2016you,yolo_v5} and DeepSORT~\cite{wojke2017simple} to detect and track pedestrians in the recorded images (processed into a panorama) from the recorded observations of the robot~\cite{yolo_deepsort}, and project these detections in 3D using the depth and scale estimates~\cite{niwa2022spatio} obtained from the ExAug perception module, as shown in Fig.~\ref{f:pose_est}. We use these position estimates for each of the 4000+ humans encountered in the training dataset to define the interaction loss. 
$J_\text{int}$ is designed to minimize the distance between human and robot trajectories as $J_\text{int}(\rho) = \mbox{min}_i \{\left|r_{t+i} - h_{t+i}\right|\}$ where $\gR=\{r_{t+1}, r_{t+2}, \ldots r_{t+N_s}\}$ and $\gH=\{h_{t+1}, h_{t+2}, \ldots h_{t+N_s}\}$ are the robot and human trajectories estimated by same approach in Section~\ref{sec:soc_nav}. Similar to $J_\text{ps}$, we only penalize the smallest $\left|r_i - h_i\right|$ by giving min formulation to better capture the desired interaction behavior.
%
%\begin{equation}
%    J_\text{int}(\rho) = \mbox{min}_i \{\left|r_{t+i} - h_{t+i}\right|\},
%    \label{eq:j_interaction}
%\end{equation}
%
%where $\gR=\{r_{t+1}, r_{t+2}, \ldots r_{t+N_s}\}$ and $\gH=\{h_{t+1}, h_{t+2}, \ldots h_{t+N_s}\}$ are the robot and human trajectories estimated by same approach in Section~\ref{sec:soc_nav}. 
%$r_i$ is robot's position estimate at time $i$ on top-down 2D plane, which is calculated by integrating velocity commands $\{v_i, \omega_i \}_{i=1 \ldots N_s}$. $h_i$ is the \emph{nearest} non-stationary pedestrian's position in the robot's local frame at time $i$. 
%Intuitively, this objective denotes the closest distance between the robot and pedestrian during their interaction, and optimizing Eqn.~\ref{eq:objective} would result in the best parameter $\rho$ that minimizes this closest distance of interaction. Similar to the design of $J_\text{ps}$, we only penalize the smallest $\left|r_i - h_i\right|$ by giving min formulation of Eqn.~\ref{eq:j_interaction} to better capture the desired interaction behavior.
%Please refer to the {\colornewparts{appendix Section X}} for relevant implementation and training details.
%

%%SL.7.14: could probably shorten this a lot and move much of it to appendix
%%NH.7.15: I shortened. I removed redundant phrases.
%%SL.2.28: use sentence case for paragraph headings
\vspace{1mm}
\noindent
{\colornewparts
\textbf{Help-and-rescue module:}
To make the data collection process as seamless and autonomous as possible, we designed a pipeline for autonomous recovery from collisions and remote help in case of irrecoverable collisions for the challenging obstacles. %(e.g., that may be shorter than the camera height, made of glass etc.). 
%When a collision is detected by the robot's collision detector sensor, an automatic backup maneuver is executed. This maneuver drives the robot away from the obstacle for a short distance along the normal vector corresponding to the point of contact. This allows the robot to automatically recover from 70\% of the simple collisions where the robot accidentally runs into challenging obstacles (e.g., that may be shorter than the camera height, made of glass etc.). Complete autonomy, however, may not be possible to achieve. The robot may drive itself into a convex hull of multiple obstacles, leading to repeated collisions, or get it's wheels stuck (e.g., on an air vent) and be unable to rescue itself. 
%%SL.3.1: We should really make sure that any claims about being "autonomous" in the paper are scoped carefully, to make sure we are not claiming anywhere that it is "fully" autonomous
We built a messaging and remote teleoperation interface, where the robot sends a signal to a remote operator when in need of remote teleoperation. }%help over the internet, and the remote operator can take control of the robot over the internet to drive the robot out of the tricky situation.
%(Fig.~\ref{f:rescue}). This simple interface also enables fleet operations, where a single operator can monitor and assist a large fleet of robots to enable data collection at scale. This module allows the robot to recover from 95\% of the more difficult situations, with only a handful cases requiring a physical intervention. Please refer to our website for relevant implementation details.
%
%
%\begin{figure}[t]
%  \begin{center}
%      \includegraphics[width=0.99\hsize]{fig/rescue_system_v3.jpeg}
%  \end{center}
%      \vspace*{0mm}
%	\caption{\small {\bf Help-and-rescue module.} The robot messages the operator for assistance when stuck (left), and can be rescued remotely over the internet using our web interface (right).}
%  \label{f:rescue}
%  \vspace*{-1.5em}
%\end{figure}
%
%%SL.7.14: could probably shorten this a lot and move much of it to appendix
%%NH.7.15: I moved details to the appendix.

\vspace{1mm}
\noindent
{\colornewparts
\textbf{Long-term anchors:} 
To overcome the limitation of localization in the repetitive environments, we place AR tags~\cite{ros_armarker} throughout the environment at approximately 10 meters apart. Since these tags are located at fixed anchor locations, we can use their coarse positions to find the corresponding nodes of our topological graph. Please see the supplemental material on our project page for more information about Help-and-rescue module and Long-term anchors. }
%to., reducing perceptual aliasing in long-term localization. Please see the supplemental material on our project page for more information. 

\vspace{1mm}
\noindent
\textbf{Continual learning:} 
As the data collection system is deployed, it may encounter novel challenges---such as varying environmental lighting throughout the day, new obstacles in the environment etc.---and it must adapt the learned data collection behavior to these changes. To achieve this, our system adopts a continual learning approach, where training data at the end of each day of deployment is used to \emph{fine-tune} the data collection policy to incorporate new experience. We also use data collected across multiple days, and times of day, to augment the fine-tuning data by chaining diverse trajectory segments between the same subgoals~\cite{chebotar2021actionable}. This allows us to effectively incorporate experience over multiple days of deployment, while also improving robustness to variations across different days and times of day. %Please see our supplemental materials for more information on trajectory chaining.
%Please see Appendix \ref{sec:app_chain} for more information on trajectory chaining.
%
\subsection{Data collection}
%% DS: Need to mention what all types of data the dataset contains, how the ground truth position labels are computed (localization with long term anchors + integrated velocities), how the pedestrian labels are computed, 
%
We use the above \DataName system with the data collection policy $\pi_{\rho}$ to autonomously collect over 75 hours of robot navigation data in 5 diverse human-occupied environments, capturing over 4000 rich interactions with humans. We describes the robotic system used for data collection environment setup, as well as key characteristics of the dataset. 
                                          
\begin{figure}[t]
  \begin{center}
      \includegraphics[width=0.99\hsize]{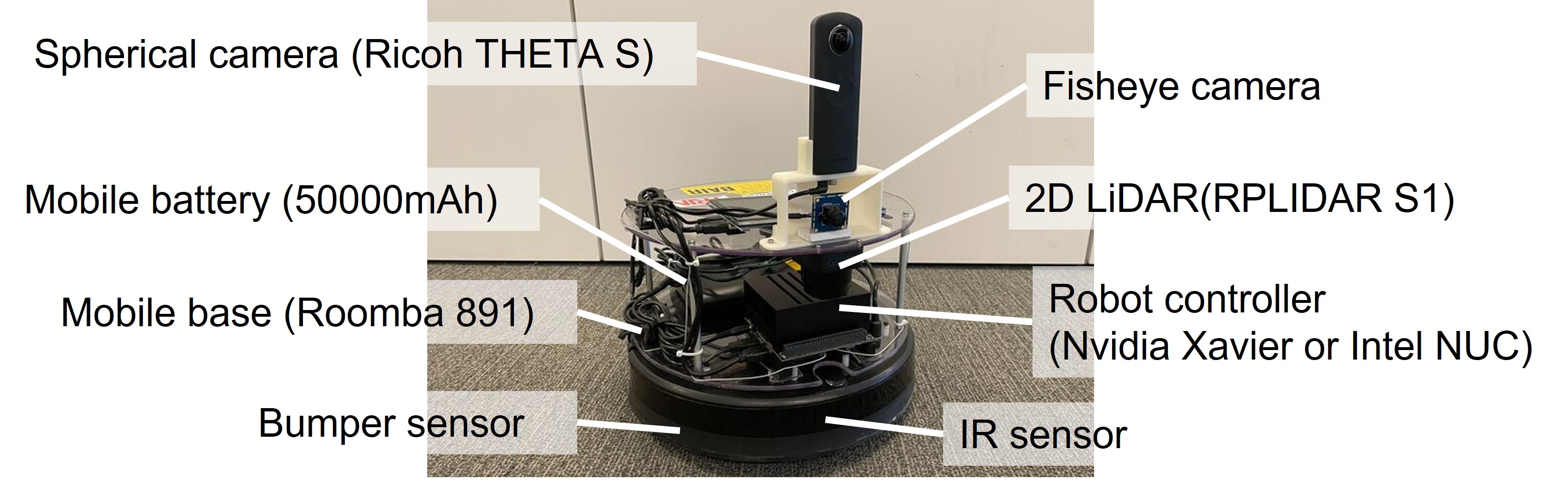}
  \end{center}
      \vspace*{0mm}
	\caption{\small {\bf Data collection platform.} We collect spherical and fisheye RGB images, 2D LiDAR, global odometry (using long-term visual anchors), and bumper signals using our robotic system.}
  \label{f:vizbot}
  \vspace*{-1.5em}
\end{figure}

\vspace{1mm}
\noindent
\textbf{Robot and Environment Setup:} 
%
%\subsection{Robot and Environment Setup}
%\label{sec:robot}
%
Figure~\ref{f:vizbot} shows an overview of our platform, built on top of an iRobot Roomba base~\cite{niwa2022spatio, ros_roomba}. The robot is equipped with two visual sensors (a spherical camera, and a $170^\circ$ wide-angle RGB camera), and a 2D LiDAR. We use two identical data collection robots with identical sensors, equipped with different onboard computers: an NVIDIA Jetson Xavier AGX, and an Intel i5 NUC, with all computation run onboard without a dedicated GPU. Our system commands angular and linear velocity commands to the base, and has access to the bumper collision sensor for triggering our help-and-rescue module.

During deployment, we instrument the environment with $N_\text{AR}$ AR tags to coarsely define the robot's route for data collection (approximately 10 m apart), and collect an example trajectory by teleoperation. This example trajectory is subsampled at a fixed frame rate of 0.5 fps to generate a topological graph of the environment. Additionally, we associate each AR tag with neighboring image nodes by collecting their ID and relative pose estimates in the robot's local frame. Starting with a \emph{base} control policy that does not encourage human interactions, \DataName system autonomously collects data that is used to train a new control policy that can interact with humans (Section~\ref{sec:interactions}) and can improve with increasing environmental experience (Section~\ref{sec:data_collection}). Please see our supplemental materials for further details.

\noindent \textbf{Implementation details:} {\colornewparts We use the same hyperparameters and architecture for training $\pi_\theta$ and $\pi_{\rho}$.} Following ExAug~\cite{hirose2022exaug}, we set the control horizon $N_s=8$ and the past observations $N_p=5$ (see Eqn.~\ref{eq:control_policy}). For pedestrian detection and tracking, we use the spherical camera on the robot to allow detection and interactions with pedestrians behind it. For the trajectory chaining procedure described in Section~\ref{sec:data_collection}, we merge multiple trajectories across several days from an environment to enhance robustness to visual distractors. We use a batch size of 80, with the training pair (past observations and subgoal images) sampled from the same trajectory for one half of the batch, and the pair coming from different trajectories in the other half of the batch. We empirically set the weights $w_i=1.5$, $w_{cp}=10.0$, and $w_{ps}=100.0$ for each objective, after analyzing closed-loop navigation performance. 

For training $\pi_\theta$, we pre-train $f_{\psi}$ with $\alpha = N_s-1$ and $\beta = N_s$ by minimizing the MSE loss using supervised learning and frozen $f_{\psi}$ while training $\pi_\theta$. We calculate the gradient of $\pi_\theta$ by back-propagation via $f_{\psi}$ for updating $\pi_\theta$. {\colornewparts To train a more accurate predictive model, we generate the human and robot trajectories by social force model~\cite{helbing1995social} and mix them with our real data in the batch.} One half of the batch is from our real dataset and the other half of the batch is from the social force model. Please see our supplemental materials for more information on the simulation data. Following \cite{Tsoi_2021_Sean_EP}, we set the personal space {\colornewparts $\mathcal{R}_h$} as 0.45 and the robot radius ${\colornewparts \mathcal{R}_r}$ as 0.25 including a small margin.
All other hyperparameters are replicated from ExAug~\cite{hirose2022exaug}.

\begin{figure}[t]
  \begin{center}
      \includegraphics[width=0.99\hsize]{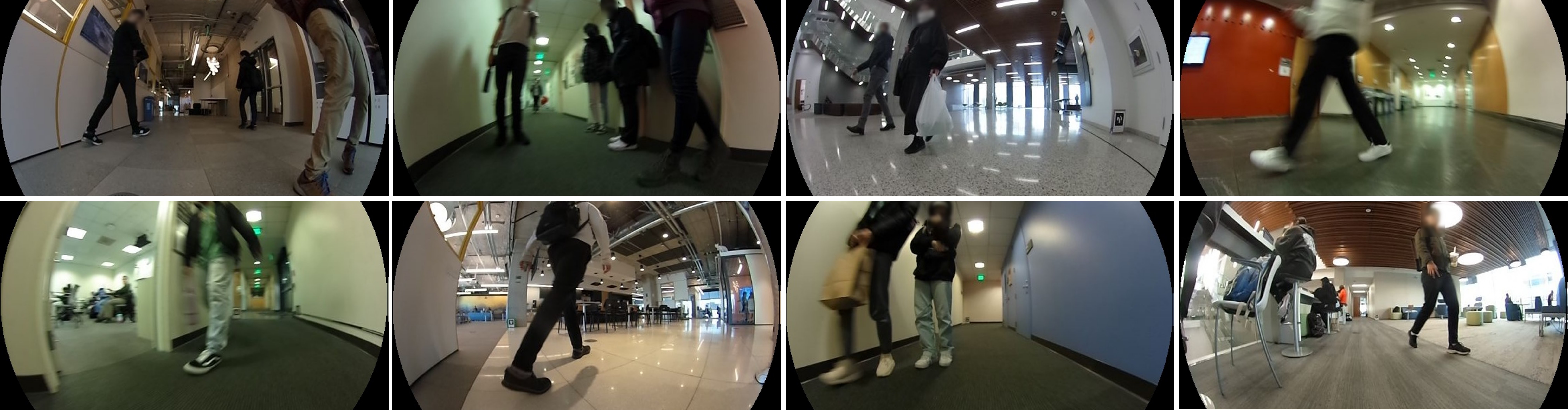}
  \end{center}
      \vspace*{0mm}
	\caption{\small {\bf Example scenes from the \DataName Dataset.} We collected our dataset in 5 different environments, spanning over 75 hours of data collection and 4000 rich human interactions, containing raw visual observations (cropped spherical images shown here).}
  \label{f:dataset}
  \vspace*{-1.5em}
\end{figure}

\vspace{1mm}
\noindent
\textbf{Dataset Characteristics:}
%
%\subsection{Dataset Characteristics}
%\label{sec:dataset}
%
We collected the \DataName dataset over the course of 24 days in 5 diverse environments, spread across 3 university buildings. The dataset spans 75 hours and 58 kilometers of autonomous robot navigation trajectories, containing over 4000 interactions with humans. The dataset includes visual observations (spherical and fisheye), 2D LiDAR scans, velocity information, and collision signals from the bumper. Figure~\ref{f:dataset} shows example images of rich human-robot interactions captured in our dataset. 

To evaluate the efficacy of the proposed interaction objective $J_\text{int}$ (Section~\ref{sec:interactions}), our dataset contains two equal subsets: the \emph{interaction-enriched dataset} corresponding to data collected by the collection policy with interaction objective ($w_i=1.5$), and the \emph{na\"ive dataset} collected without ($w_i=0$). We have released this dataset publicly on our project page.

%\noindent \textbf{Quantifying human interactions:}
%To understand the effect of the proposed interaction loss on the quality of the data collected, we conduct controlled experiments with 5 human participants tasked with interacting with the data collection system running two different collection policies: one that encourages interactions and another that does not.
%
%While quantifying the amount of human interaction is a challenging problem by itself~\cite{wang2022metrics}, we propose three metrics that coarsely capture these interactions: (i) the mean distance of the robot to an observed pedestrian, (ii) bounding box area (in sq. pixels) of the observed pedestrian, as detected by an object detector~\cite{redmon2016you}, and (iii) the offset (in pixels) of the observed pedestrian from the center of the robot's frame (e.g., this would correspond to the visual servoing error for a follower robot~\cite{giesbrecht2009vision}). Table~\ref{tab:ev1} shows the results of this evaluation on the two subsets of our dataset. We observe that the interaction-enriched dataset results in collection policies that drive closer to the pedestrians, and captures more prominent interactions.

\section{Evaluation}
We design our experiments to evaluate the socially compliant control policy $\pi_{\theta}$ with our proposed objectives $J_\text{cp}$ and $J_\text{ps}$, as well as the proposed interaction-enriched dataset collected by our autonomous data collection system. Specifically, we study the following questions:
\begin{enumerate}[label={\bf Q\arabic{*}.}, leftmargin=2.8\parindent]
    \item Does our proposed objective lead to better socially unobtrusive behavior?
    %%SL.7.13: What do you mean by "engagement"? Our proposed objective is meant to avoid preturbing the human, but perhaps this question is a holdover from back when we were studying *maximizing* engagement? I would also be careful with the phrase "socially consistent" as it's very overloaded and might invite lots of criticism from human-robot interaction folks about how there is a lot more to social interaction than just avoiding humans. Maybe unobtrusive or somethign like that might be a better word
    %%NH.7.15: DONE
    \item Does our proposed data collection system lead to more interactions, and does this in turn lead to better predictive models of pedestrians?
    \item How does the navigation capabilities of our policy improve over the course of collecting our dataset?
    %\item How does the performance of the data collection system evolve with increasing experience?     
    %%SL.7.13: It's unclear what the relevance of this question is now, since the motivation is more about the final nav policy, but perhaps we can rephrase this question to be more self-explanatory? Or add more discussion at the top of this section so that this is less surprising?
    %%NH.7.15: I revised.
\end{enumerate}
\subsection{Socially Compliant Navigation}

Towards answering \textbf{Q1}, we train two different policies with and without our proposed objectives $J_\text{cp}$ and $J_\text{ps}$. Here, the control policy without $J_\text{cp}$ and $J_\text{ps}$ corresponds to the most relevant baseline method, ExAug~\cite{hirose2022exaug}. In addition, we train different social navigation policy on the na\"ive dataset without the proposed interaction objective.
%%SL.7.13: Explaining the baselines is really important, maybe we should have a dedicated paragraph where we explain this clearly along these lines: We compare our method to baselines that evaluate both the utility of our proposed objective and the importance of our dataset, which is collected with an explicit interaction-seeking policy. For the former, we compare to baselines that [do whatever]. For the latter, we compare to a baseline that uses the same objective as our method, but is trained on [whatever]. (and it would also be good to *name* these baselines, mention the name when we introduce them, and refer to them by that name later for clarity)
{\colornewparts We conduct fifteen experiments using the real robot across a few days (five experiments in each three difference real environments). In these experiments, we use goal images which were collected over two months ago to evaluate the robustness of the policies to environmental changes.} The distance between the start and goal positions ranges from 13.0 to 37.8 meters, which is considered relatively long for vision-based navigation in indoor settings. In order to ensure equivalent experimental conditions, we request during the evaluation that the pedestrians navigate around the robot, creating similar interaction scenarios for each control policy. If the robot collides with a pedestrian or obstacle, we request the pedestrian to distance themselves from the robot's perimeter, and we allow the robot to continue navigation.
%And, we asked the pedestrians to do not stop navigation after collisions, if the robot continues to move.
%%SL.7.13: I'm not sure what this last sentence means, perhaps we can rephrase it?
%%NH.7.15: I modified the explanation.
%
\begin{table}[t]
  \begin{center}
  \vspace{0mm}
  \caption{{\small {\bf Closed-loop Evaluation of trained control policies.} {\colornewparts We evaluate the performance on different training datasets with and without $J_\text{int}$. We also evaluate the performance with and without objectives $J_\text{cp}$ and $J_\text{ps}$ during training. The policy without $J_\text{cp}$ and $J_\text{ps}$ is ExAug.}}}  
  \label{tab:ev2}  
  \resizebox{1.0\columnwidth}{!}{
  \begin{tabular}{llccccccc} \toprule 
   Method & Training dataset & GR $\uparrow$ & SPL $\uparrow$ & STL $\uparrow$ & CP $\downarrow$ [\#] & CO $\downarrow$ [\#] & PSV [s] $\downarrow$ \\ \midrule
   ExAug~\cite{hirose2022exaug} & with $J_\text{int}$ (ours) & 0.800 & 0.692 & 0.595 & 20 & 6 & 85.248 \\  
   Ours & no $J_\text{int}$ (baseline) & 0.667 & 0.517 & 0.365 & 8 & 11 & 84.915 \\
   Ours & with $J_\text{int}$ (ours) & {\bf 1.000} & {\bf 0.888} & {\bf 0.692} & {\bf 1} & {\bf 2} & {\bf 57.609} \\ \bottomrule
  \end{tabular}
  }
  \end{center}
  %\caption{{\small {\bf Closed-loop Evaluation of trained control policies.} We find a policy trained with the interaction-enriched dataset results in less collisions and personal space violations for social navigation and enables the robot to reach the goal position at higher accuracy.}}  
   \vspace{-1em}
\end{table}
\begin{figure}[t]
  \centering
    \includegraphics[width=0.92\linewidth]{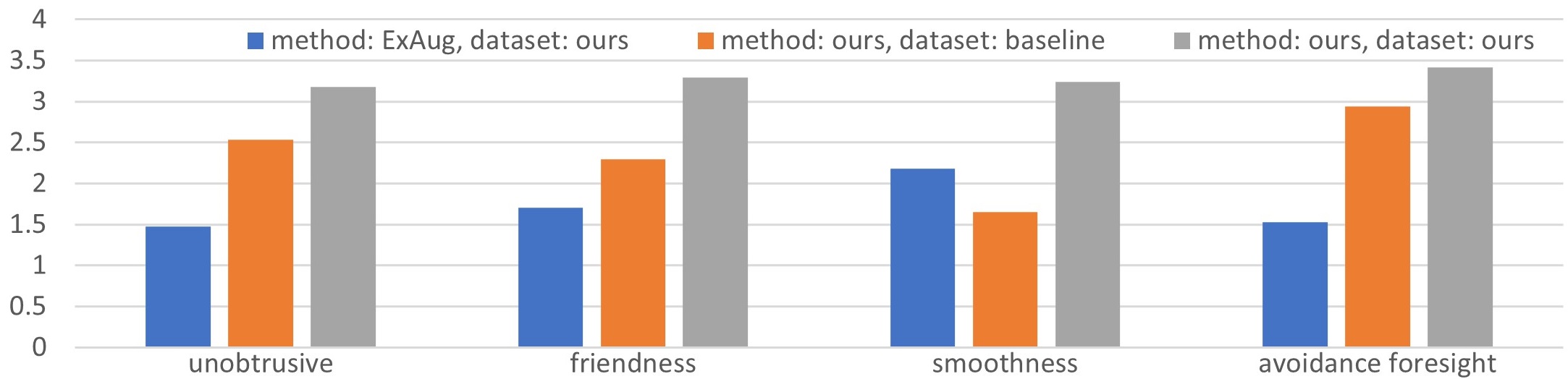}
	\caption{\small {\bf Evaluation of socialness by human rating.} The values are average of scores from 17 subjects. Larger is better in all ratings.}
	\label{f:human_rate}
    \vspace{-1.5em}
\end{figure}

Table~\ref{tab:ev2} presents the comparison of our method to the above baselines along several metrics: Goal arrival Rate (GR), Success weighted by Path Length (SPL) \cite{anderson2018evaluation}, Success weighted by Time Length (STL) \cite{francis2023principles}, Collision count for Pedestrians (CP), Collision count for static Objects (CO), and Personal Space Violation duration (PSV). Our control policy trained on our proposed dataset with $J_\text{cp}$ and $J_\text{ps}$ shows a clear improvement over ExAug and the control policy trained on the na\"ive dataset without $J_\text{int}$ in all metrics. In particular, our method decreases the collision counts for pedestrians by more than 80$\%$, reduces PSV by over 30$\%$, and successfully leads the robot to the goal position. 

{\colornewparts Furthermore, we conduct a user study to evaluate the robot's behavior in real-world environments. We recruited 17 subjects from a university campus, encompassing diverse genders, races, and backgrounds; however, there was a bias with 80\% being students. We conduct 3 navigation experiments by three different control policies in Table II for each subject (51 experiments in total).  We ask them to walk around the robot without explaining which control policy we are running, and we have them evaluate the social compliance and smoothness of our policy between 4 and 0 (larger is better) for four questions after each experiment. Fig. 7 demonstrates the advantage of our method in human ratings across all questions.} The comparison suggests that our proposed objective improves the robot's ability to navigate unobtrusively in the presence of humans, and our proposed dataset collected via an interaction-seeking policy leads to better performance for our method.

\begin{figure}[t]
  \begin{center}
      \includegraphics[width=0.92\hsize]{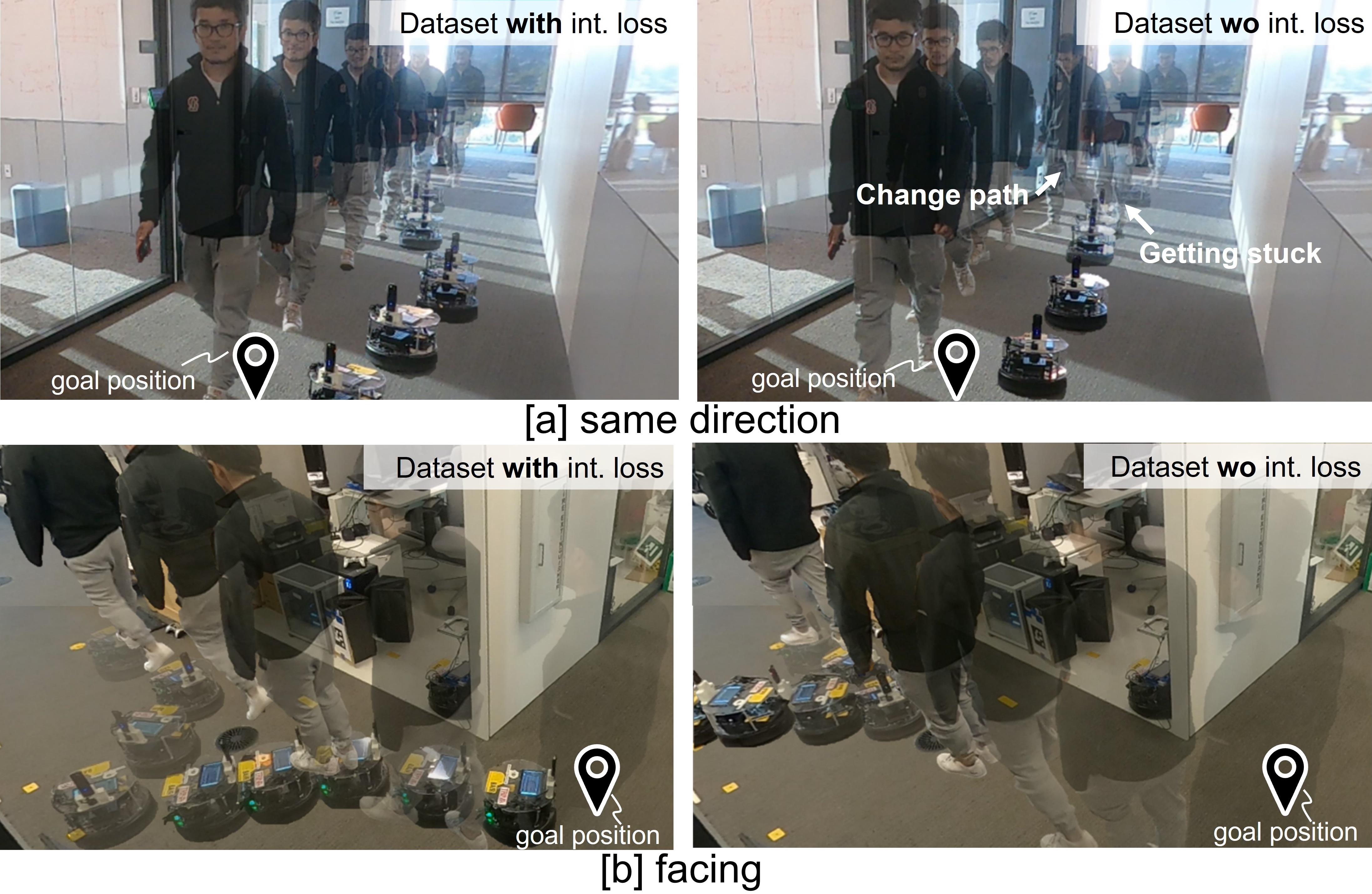}
  \end{center}
      \vspace*{-3mm}
	\caption{\small {\bf Qualitative Examples of Learned Behavior.} A social navigation policy trained on the interaction-enriched subset of \DataName (left) leads to better handling of human pedestrians while successfully reaching the goal, without intruding in their personal space. Training on the na\"ive dataset results in a conservative policy (right) that gets stuck and collides with pedestrian.}
  \label{f:visualization}
  \vspace*{0mm}
\end{figure}

In Fig.~\ref{f:visualization}, we qualitatively observe the robot's behavior to be significantly more ``compliant'' when trained with the interaction-enriched dataset (left). Even in the narrow corridors, our control policy makes space for the pedestrians while still maintaining clearance from the walls. The control policy trained on the na\"ive dataset does not take avoidance action when a pedestrian approaches the robot, so the robot often violates personal space, collides with the pedestrian (top right), or fails to reach the goal (bottom right).

\begin{table}[t]
  \begin{center}
  \vspace{0mm}
  \caption{{\small {\bf Training a Pedestrian Dynamics Model.} {\colornewparts We evaluate performance on different training datasets with and without $J_\text{int}$, along with the effect of data presence from the social forces model.}}}  
  \label{tab:ev_traj_est}  
  %\caption{{\small {\bf Training a Pedestrian Dynamics Model.} Training on the interaction-enriched dataset leads to better predicted trajectories.}}    
  \resizebox{0.75\columnwidth}{!}{
  \begin{tabular}{lccc} \toprule 
    Real dataset & {\colornewparts Social force model}~\cite{helbing1995social} & MSE $\downarrow$ & Cosine $\uparrow$ \\ \midrule
    -- & \checkmark & 0.0147 & 0.792 \\
    no $J_\text{int}$ (baseline) & & 0.0099 & 0.852 \\
    no $J_\text{int}$ (baseline) & \checkmark & 0.0095 & 0.856 \\    
    with $J_\text{int}$ (ours) & & 0.0084 & 0.872\\
    with $J_\text{int}$ (ours) & \checkmark & {\bf 0.0083} & {\bf 0.876}\\ \bottomrule    
  \end{tabular}
  }
  \end{center}
   \vspace{-1.7em}
\end{table}
\subsection{The Value of Interaction-Rich Data}
%\subsection{Modeling Pedestrian Dynamics}
\noindent \textbf{Modeling Pedestrian Dynamics:} While the previous evaluation studies the end-to-end performance of our system, in the next experiment we specifically examine the pedestrian prediction model at the core of our method, and how its predictive accuracy changes based on the composition of the training dataset. Our aim is to understand whether our proposed interaction-seeking data collection scheme actually leads to more accurate pedestrian prediction models.
%%SL.7.13: Feels like we need more motivation for why we're studying this, maybe we can make the first sentence below more explicit, adding something like: While the previous evaluation studies the end-to-end performance of our system, in the next experiment we specifically examine the pedestrian prediction model at the core of our method, and how its predictive accuracy changes based on the composition of the training dataset. Our aim is to understand whether our proposed interaction-seeking data collection scheme actually leads to more accurate pedestrian prediction models.
%%NH.7.15: DONE
%Modeling and forecasting pedestrian behavior is an important milestone towards robotic systems that can interact with humans, and has been studied in a wide variety of applications~\cite{herman2021pedestrian, kiss2021probabilistic, martin2021jrdb}. 
For \textbf{Q2}, we train the predictive model $f_{\psi}$ on a combination of three datasets: the interaction-enriched dataset, the na\"ive dataset, and the simulation dataset from social force model~\cite{helbing1995social}. We report the mean squared error, to capture how close each predicted point is to the true future positions, and the cosine similarity score, that measures the alignment between the vectors corresponding to the predicted and true positions (a scale-invariant metric proposed in GNM~\cite{shah2022gnm}). 

\begin{figure}[t]
  \begin{center}
      \includegraphics[width=0.92\hsize]{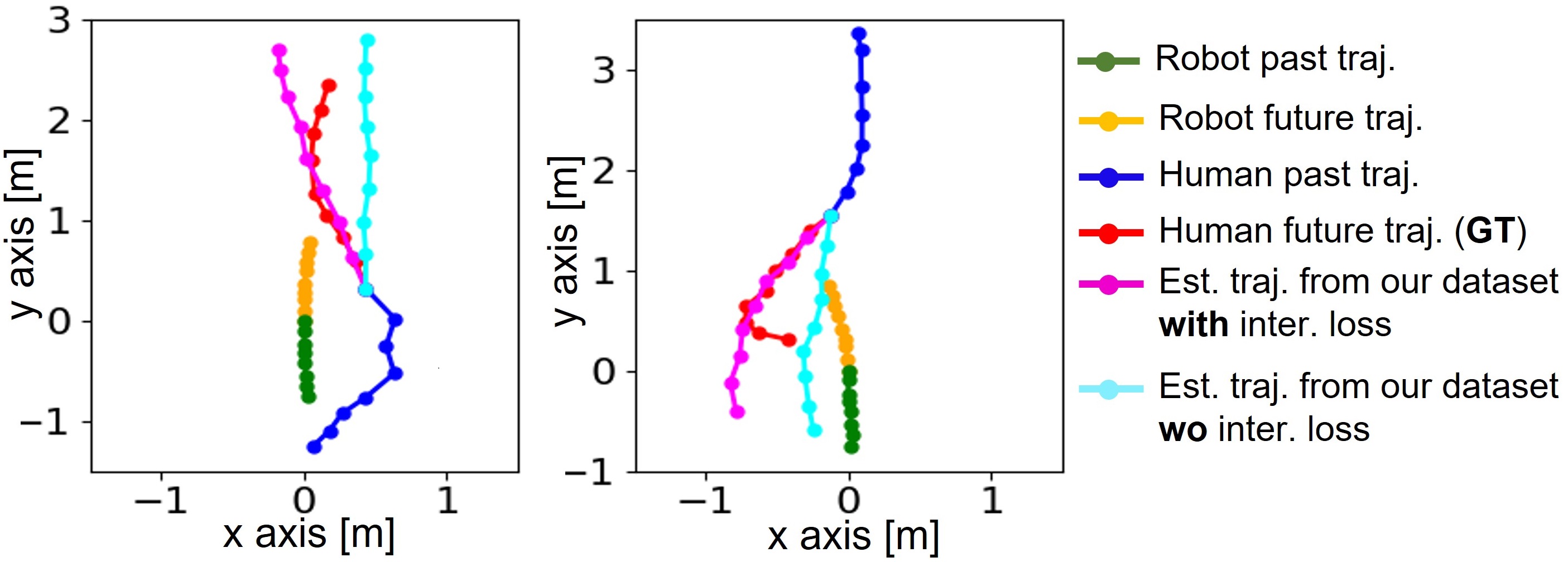}
  \end{center}
      \vspace*{-3mm}
	\caption{\small \textbf{Examples of using the learned human dynamics model to predict future human positions}, conditioned on past human positions (blue), past robot positions (green), and future robot plans (yellow). Training with the interaction-enriched dataset (magenta) leads to better predictions than the na\"ive dataset (cyan).}
%%SL.2.28: I think this figure is a bit hard to interpret, can we add more discussion about what readers should take away from this figure, and maybe in the process better explain how to interpret it?
  \label{f:est_traj}
  \vspace{-1.5em}
\end{figure}

Table~\ref{tab:ev_traj_est} shows the evaluation results of the predictive model. We find that a predictive model trained with the interaction-enriched dataset leads to better predictions, both in terms of the direction and scale, suggesting that the proposed objective indeed allows better prediction of future human behavior. In addition, its performance is much better than the trained model solely from the simulator. Since the simulation dataset from social force model can help to improve the predictive performance by mixing with our real dataset in training, we use these models for training the socially compliant control policy in Table~\ref{tab:ev2}. Fig.~\ref{f:est_traj} illustrates the predictive model in action for two example interactions. Estimated trajectories of our predictive model trained on our dataset with $J_\text{int}$ (magenta) coincide well with the ground truth pedestrians trajectories (red), different from the estimated trajectories trained on the na\"ive dataset (cyan).
%Table~\ref{tab:ev_traj_est} shows the evaluation results of the predictive model trained on combinations with datasets with enriched and non-enriched human and robot interaction and the dataset from social force mode. We find that a predictive model trained with the interaction-enriched dataset leads to better predictions, both in terms of the direction and scale, suggesting that the proposed objective indeed allows better prediction of future human behavior. Fig.~\ref{f:est_traj} illustrates the predictive model in action for two example interactions.

Moreover, to investigate the effect of the proposed interaction loss on the quality of the data collected, we conduct controlled experiments with 5 human participants tasked with interacting with the data collection system running two different collection policies: one that encourages interactions and another that does not. While quantifying the amount of human interaction is a challenging problem by itself~\cite{wang2022metrics}, we propose three metrics that coarsely capture these interactions: (i) the mean distance of the robot to an observed pedestrian, (ii) bounding box area (in sq. pixels) of the observed pedestrian, as detected by an object detector~\cite{redmon2016you}, and (iii) the offset (in pixels) of the observed pedestrian from the center of the robot's frame (e.g., this would correspond to the visual servoing error for a follower robot~\cite{giesbrecht2009vision}). Table~\ref{tab:ev1} shows the results of this evaluation on the two subsets of our dataset. We observe the explicit difference, which results in better predictive model and socially compliant control policies.

\begin{table}[t]
  \begin{center}
  \caption{{\small {\bf Evaluation of the interaction objective.} A policy trained on the interaction-enriched dataset (with $J_\text{int}$) drives closer to the pedestrians, and captures more prominent interactions.}}  
  \resizebox{0.8\columnwidth}{!}{
  \begin{tabular}{lccc} \toprule 
    Dataset &  Distance [m] $\downarrow$ & Area [px$^2$] $\uparrow$ & Offset [px] $\downarrow$ \\ \midrule
    no Jint (baseline) & 2.67 & 0.99 $\times$ 10$^4$ & 128.51 \\
    with $J_\text{int}$ (ours) & {\bf 2.43}  & {\bf 1.40 $\times$ 10$^4$} & {\bf 98.55}\\ \bottomrule
  \end{tabular}
  }
  \label{tab:ev1}
  \end{center}
  \vspace*{-1em}
\end{table}

\begin{figure}[t]
  \begin{center}
      \includegraphics[width=0.92\hsize]{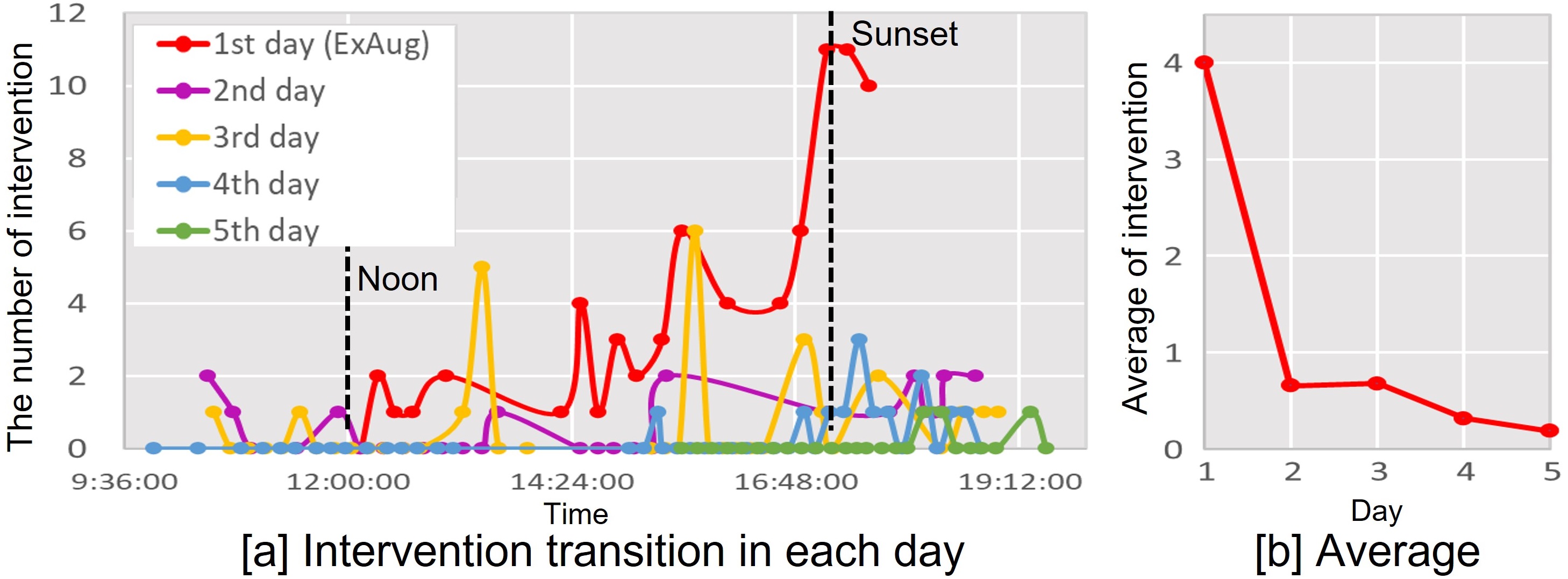}
  \end{center}
    \vspace*{-3mm}  
	\caption{\small {\bf Self-Improvement with Continual Learning.} \DataName can improve with increasing experience, a 95.5\% reduction in interventions over the course of 5 days (right). While the performance varies across times of day due to variability in appearances (left, red), continual learning enables our system to eventually learn consistent collision-free behavior across the day (green).}
  \label{f:graph_cl}
\vspace{-1.3em}
\end{figure}

\noindent \textbf{Continual Learning with the \DataName System:} Lastly, we evaluate how the navigation capabilities of the robotic policy improve over the course of collecting our dataset. While this experiment does not directly evaluate the robot's ability to interact with humans, it does show how our data collection system can enable autonomous improvement, validating the scalability of our data gathering approach for \textbf{Q3}.
We deploy \DataName system to operate autonomously, with occasional remote assistance, to collect data through the environment. At the end of a collection day, this data is used to \emph{fine-tune} the policy to incorporate the new experience (as described in Sec.~\ref{sec:data_collection}) .
Figure~\ref{f:graph_cl}(a) shows the average number of remote interventions requested by the data collection system during different times of the day. We notice that at the start, the variability in environmental lighting is significant and the initialized model (red) performs significantly worse as the day progresses. However, \DataName system is able to quickly incorporate this new experience and improve it's performance in subsequent data collection days, requiring fewer interventions each time. Over the course of multiple days (b), our system learns near-perfect autonomous navigation in the challenging indoor environment with dynamic obstacles, requesting an average of 0.18 interventions per a 10 minute trajectory, representing a 95.5\% improvement over the day 1 baseline.
\section{Discussion}
In this paper, we proposed a method for training the \MethodName policy for vision-based navigation to build the socially unobstrusive navigation system. In training \MethodName policy, we introduced novel objectives using the predictive model of the pedestrians' future trajectories to suppress the counterfactual perturbation from the intended human trajctories. To obtain an accurate predictive model for a better \MethodName policy, we proposed the \DataName system, a scalable data collection system, to autonomously collect a dataset with enriched human-robot interactions. \DataName system has the data collection control policy to interact with the pedestrians while collecting the dataset.
%, in addition to three key components: (i) a help-and-rescue module, (ii) a long-term localization module, and (iii) a continual learning framework.
%In this paper, we proposed a method for training a socially compliant robotic navigation controller, by combining a novel counterfactual perturbation objective with a data collection system that incentivizes interaction to collect data that is well suited for learning predictive models of pedestrian interactions. The \MethodName system is modular, can be built around an existing vision-based navigation system, and has three key components: (i) a help-and-rescue module for collision recovery and seeking remote assistance, (ii) a long-term localization module that uses AR markers placed around the environment for coarse guidance, and (iii) a continual learning framework that allows the system's performance to improve with increasing experience.
%%SL.7.13: I think the above sentence is out of place in the discussion section, since the paper is now not just about the data collection but about the entire system, perhaps the above can be replaced with something that more logically follows from the first sentence?
%%NH.7.15: Thank you. I modified.
We used this data collection system to collect the \DataName dataset: the publicly available dataset of visual navigation around humans, spanning over 75 hours of data collected in 5 different environments and comprising over 4000 rich human-robot interactions. Our experiments show that policies trained on the collected dataset enables the real robot to navigate with the socially unobtrusive behavior.
%Our experiments show that collecting a large dataset with an interaction enriching objective can lead to improvements in learning \MethodName policy suppressing the counterfactual perturbation and pedestrian behavior modeling in real-world environments.
%%SL.7.13: This bit is reasonable, but seems to focus exclusively on the dataset, perhaps we can revise this paragraph to provide a more balanced treatment that focuses on both the method and the dataset?
%%NH.7.15: I modified a bit.

{\colornewparts Our \MethodName policy, when trained on a dataset with enriched human-robot interactions, still has some limitations. Our current system only learns simple social interactions such as avoiding a pedestrian's personal space and giving way to the pedestrians by considering the closest pedestrians' behavior. To understand more complex scenes, we will need to incorporate better objectives accounting for multiple pedestrians and their grouping in the data collection and deployment policies. } 
%Additionally, our autonomous robot is restricted to operate at slow speeds (capped at 0.4 m/s) to limit damage due to policy errors in safety-critical environments with humans; however, this inherently limits the robot behavior. %When a pedestrian is approaching the robot at a casual pace, even if the robot takes actions to avoid the pedestrian, it may not be fast enough.  

We believe that \DataName opens up many exciting avenues for socially compliant navigation systems in human inhabited spaces. 
%We believe that \MethodName is the opens up many exciting avenues for autonomous collection systems in human inhabited spaces. 
The possibility of scaling such as system to new environments and platforms and objectives is promising. {\colornewparts The limitation of the \DataName dataset we present is that it lacks complex scenes that include groups of multiple pedestrians. Also, the environments in the dataset are limited to office buildings.}
%%SL.7.13: It's quite important to end the discussion on a bit more of a high note by discussing future work. Future work often follows naturally from limitations, but it's good for the future work to be inspiring and get readers to think that this work will lead to lots of great stuff in the future.
%%[TODO]Furthermore
%
%Our instantiation of the autonomous data collection principle does have some limitations. Most prominently, our autonomous robot is restricted to operate at slow speeds (capped at 0.4 m/s) to limit damage due to policy errors in safety-critical environments with humans; however, this inherently limits the quality of interactions observed by the robot. Humans walking at a casual pace are much faster than the robot, and tend to not interact closely with the robot, for instance, to give right-of-way, or open the door etc. The trade-off between capturing complex social interactions, and staying safe around humans, needs to be studied carefully to result in more interactive systems that demonstrate complex, cooperative behaviors. Additionally, our current system only learns simple social interactions such as avoiding a pedestrian's personal space and giving way to the pedestrians; learning richer social interactions will need better objectives incorporated more tightly in the data collection and deployment policies.

\section*{Acknowledgments}
This research was supported by Berkeley DeepDrive at the University of California, Berkeley, and Toyota Motor North America. Additionally, partial support for this research was provided by ARL DCIST CRA W911NF-17-2-0181. The authors would like to express their gratitude to Marwa Abdulhai, Qiyang Li, Manan Tomar, Mitsuhiko Nakamoto, Roxana Infante, Ami Katagiri, Katie Kang, Zheyuan Hu, Oier Mees, Jakub Grudzien Kuba, Pranav Atreya, Isadora White, Zhiyuan Zhou, Anjali Thakrar, Niclas Joswig, Kyle Stachowicz, and Catherine Glossop for their valuable assistance in evaluating the \MethodName.

We consulted the Committee for Protection of Human Subjects at our home university (UC Berkeley) and it was determined that the study does not meet the definition of research with human subjects set forth in Federal Regulations at 45 CFR 46.102.

\bibliographystyle{IEEEtran}
\vskip-\parskip
\begingroup
\footnotesize
%\scriptsize
\balance
\bibliography{egbib_full.bib}

\begin{thebibliography}{10}
\providecommand{\url}[1]{#1}
\csname url@rmstyle\endcsname
\providecommand{\newblock}{\relax}
\providecommand{\bibinfo}[2]{#2}
\providecommand\BIBentrySTDinterwordspacing{\spaceskip=0pt\relax}
\providecommand\BIBentryALTinterwordstretchfactor{4}
\providecommand\BIBentryALTinterwordspacing{\spaceskip=\fontdimen2\font plus
\BIBentryALTinterwordstretchfactor\fontdimen3\font minus
  \fontdimen4\font\relax}
\providecommand\BIBforeignlanguage[2]{{%
\expandafter\ifx\csname l@#1\endcsname\relax
\typeout{** WARNING: IEEEtran.bst: No hyphenation pattern has been}%
\typeout{** loaded for the language `#1'. Using the pattern for}%
\typeout{** the default language instead.}%
\else
\language=\csname l@#1\endcsname
\fi
#2}}

\bibitem{helbing1995social}
D.~Helbing and P.~Molnar, ``Social force model for pedestrian dynamics,''
  \emph{Physical review E}, vol.~51, no.~5, p. 4282, 1995.

\bibitem{ferrer2013robot}
G.~Ferrer, A.~Garrell, and A.~Sanfeliu, ``Robot companion: A social-force based
  approach with human awareness-navigation in crowded environments,'' in
  \emph{2013 IEEE/RSJ International Conference on Intelligent Robots and
  Systems}.\hskip 1em plus 0.5em minus 0.4em\relax IEEE, 2013, pp. 1688--1694.

\bibitem{martin2021jrdb}
R.~Martin-Martin, M.~Patel, H.~Rezatofighi, A.~Shenoi, J.~Gwak, E.~Frankel,
  A.~Sadeghian, and S.~Savarese, ``Jrdb: A dataset and benchmark of egocentric
  robot visual perception of humans in built environments,'' \emph{IEEE
  transactions on pattern analysis and machine intelligence}, 2021.

\bibitem{karnan2022socially}
H.~Karnan, A.~Nair, X.~Xiao, G.~Warnell, S.~Pirk, A.~Toshev, J.~Hart,
  J.~Biswas, and P.~Stone, ``Socially compliant navigation dataset (scand): A
  large-scale dataset of demonstrations for social navigation,'' \emph{IEEE
  Robotics and Automation Letters}, vol.~7, no.~4, pp. 11\,807--11\,814, 2022.

\bibitem{thorDataset2019}
A.~Rudenko, T.~P. Kucner, C.~S. Swaminathan, R.~T. Chadalavada, K.~O. Arras,
  and A.~J. Lilienthal, ``Th{\"o}r: Human-robot navigation data collection and
  accurate motion trajectories dataset,'' \emph{IEEE Robotics and Automation
  Letters}, vol.~5, no.~2, pp. 676--682, 2020.

\bibitem{Geiger2012CVPR}
A.~Geiger, P.~Lenz, and R.~Urtasun, ``Are we ready for autonomous driving? the
  kitti vision benchmark suite,'' in \emph{2012 IEEE conference on computer
  vision and pattern recognition}.\hskip 1em plus 0.5em minus 0.4em\relax IEEE,
  2012, pp. 3354--3361.

\bibitem{carlevaris2016university}
N.~Carlevaris-Bianco, A.~K. Ushani, and R.~M. Eustice, ``University of michigan
  north campus long-term vision and lidar dataset,'' \emph{The International
  Journal of Robotics Research}, vol.~35, no.~9, pp. 1023--1035, 2016.

\bibitem{hirose2019deep}
N.~Hirose, F.~Xia, R.~Mart{\'\i}n-Mart{\'\i}n, A.~Sadeghian, and S.~Savarese,
  ``Deep visual mpc-policy learning for navigation,'' \emph{IEEE Robotics and
  Automation Letters}, vol.~4, no.~4, pp. 3184--3191, 2019.

\bibitem{yan2020robot}
Z.~Yan, S.~Schreiberhuber, G.~Halmetschlager, T.~Duckett, M.~Vincze, and
  N.~Bellotto, ``Robot perception of static and dynamic objects with an
  autonomous floor scrubber,'' \emph{Intelligent Service Robotics}, vol.~13,
  no.~3, pp. 403--417, 2020.

\bibitem{shah2021rapid}
D.~Shah, B.~Eysenbach, N.~Rhinehart, and S.~Levine, ``Rapid exploration for
  open-world navigation with latent goal models,'' in \emph{Conference on Robot
  Learning}.\hskip 1em plus 0.5em minus 0.4em\relax PMLR, 2022, pp. 674--684.

\bibitem{nguyen2023toward}
D.~M. Nguyen, M.~Nazeri, A.~Payandeh, A.~Datar, and X.~Xiao, ``Toward
  human-like social robot navigation: A large-scale, multi-modal, social human
  navigation dataset,'' \emph{arXiv preprint arXiv:2303.14880}, 2023.

\bibitem{mavrogiannis2021core}
C.~Mavrogiannis, F.~Baldini, A.~Wang, D.~Zhao, P.~Trautman, A.~Steinfeld, and
  J.~Oh, ``Core challenges of social robot navigation: A survey,'' \emph{ACM
  Transactions on Human-Robot Interaction}, vol.~12, no.~3, pp. 1--39, 2023.

\bibitem{sisbot2007human}
E.~A. Sisbot, L.~F. Marin-Urias, R.~Alami, and T.~Simeon, ``A human aware
  mobile robot motion planner,'' \emph{IEEE Transactions on Robotics}, vol.~23,
  no.~5, pp. 874--883, 2007.

\bibitem{mumm2011human}
J.~Mumm and B.~Mutlu, ``Human-robot proxemics: physical and psychological
  distancing in human-robot interaction,'' in \emph{Proceedings of the 6th
  international conference on Human-robot interaction}, 2011, pp. 331--338.

\bibitem{mehta2016autonomous}
D.~Mehta, G.~Ferrer, and E.~Olson, ``Autonomous navigation in dynamic social
  environments using multi-policy decision making,'' in \emph{2016 IEEE/RSJ
  International Conference on Intelligent Robots and Systems (IROS)}.\hskip 1em
  plus 0.5em minus 0.4em\relax IEEE, 2016, pp. 1190--1197.

\bibitem{luber2012socially}
M.~Luber, L.~Spinello, J.~Silva, and K.~O. Arras, ``Socially-aware robot
  navigation: A learning approach,'' in \emph{2012 IEEE/RSJ International
  Conference on Intelligent Robots and Systems}.\hskip 1em plus 0.5em minus
  0.4em\relax IEEE, 2012, pp. 902--907.

\bibitem{ziebart2009planning}
B.~D. Ziebart, N.~Ratliff, G.~Gallagher, C.~Mertz, K.~Peterson, J.~A. Bagnell,
  M.~Hebert, A.~K. Dey, and S.~Srinivasa, ``Planning-based prediction for
  pedestrians,'' in \emph{2009 IEEE/RSJ International Conference on Intelligent
  Robots and Systems}.\hskip 1em plus 0.5em minus 0.4em\relax IEEE, 2009, pp.
  3931--3936.

\bibitem{bajcsy2019scalable}
A.~Bajcsy, S.~L. Herbert, D.~Fridovich-Keil, J.~F. Fisac, S.~Deglurkar, A.~D.
  Dragan, and C.~J. Tomlin, ``A scalable framework for real-time multi-robot,
  multi-human collision avoidance,'' in \emph{2019 international conference on
  robotics and automation (ICRA)}.\hskip 1em plus 0.5em minus 0.4em\relax IEEE,
  2019, pp. 936--943.

\bibitem{pfeiffer2016predicting}
M.~Pfeiffer, U.~Schwesinger, H.~Sommer, E.~Galceran, and R.~Siegwart,
  ``Predicting actions to act predictably: Cooperative partial motion planning
  with maximum entropy models,'' in \emph{2016 IEEE/RSJ International
  Conference on Intelligent Robots and Systems (IROS)}.\hskip 1em plus 0.5em
  minus 0.4em\relax IEEE, 2016, pp. 2096--2101.

\bibitem{mavrogiannis2018social}
C.~I. Mavrogiannis, W.~B. Thomason, and R.~A. Knepper, ``Social momentum: A
  framework for legible navigation in dynamic multi-agent environments,'' in
  \emph{Proceedings of the 2018 ACM/IEEE International Conference on
  Human-Robot Interaction}, 2018, pp. 361--369.

\bibitem{xiao2022learning}
X.~Xiao, T.~Zhang, K.~Choromanski, E.~Lee, A.~Francis, J.~Varley, S.~Tu,
  S.~Singh, P.~Xu, F.~Xia, \emph{et~al.}, ``Learning model predictive
  controllers with real-time attention for real-world navigation,'' \emph{arXiv
  preprint arXiv:2209.10780}, 2022.

\bibitem{truong2017approach}
X.-T. Truong and T.-D. Ngo, ``“to approach humans?”: A unified framework
  for approaching pose prediction and socially aware robot navigation,''
  \emph{IEEE Transactions on Cognitive and Developmental Systems}, vol.~10,
  no.~3, pp. 557--572, 2017.

\bibitem{chen2023social}
H.-Y. Chen, P.-H. Huang, and L.-C. Fu, ``Social crowd navigation of a mobile
  robot based on human trajectory prediction and hybrid sensing,''
  \emph{Autonomous Robots}, pp. 1--13, 2023.

\bibitem{bhaskara2023sg}
R.~Bhaskara, M.~Chiu, and A.~Bera, ``Sg-lstm: Social group lstm for robot
  navigation through dense crowds,'' \emph{arXiv preprint arXiv:2303.04320},
  2023.

\bibitem{narayanan2023ewarenet}
V.~Narayanan, B.~M. Manoghar, R.~P. RV, and A.~Bera, ``Ewarenet: Emotion-aware
  pedestrian intent prediction and adaptive spatial profile fusion for social
  robot navigation,'' in \emph{2023 IEEE International Conference on Robotics
  and Automation (ICRA)}.\hskip 1em plus 0.5em minus 0.4em\relax IEEE, 2023,
  pp. 7569--7575.

\bibitem{rosmann2017online}
C.~R{\"o}smann, M.~Oeljeklaus, F.~Hoffmann, and T.~Bertram, ``Online trajectory
  prediction and planning for social robot navigation,'' in \emph{2017 IEEE
  International Conference on Advanced Intelligent Mechatronics (AIM)}.\hskip
  1em plus 0.5em minus 0.4em\relax IEEE, 2017, pp. 1255--1260.

\bibitem{chen2018robot}
Z.~Chen, C.~Song, Y.~Yang, B.~Zhao, Y.~Hu, S.~Liu, and J.~Zhang, ``Robot
  navigation based on human trajectory prediction and multiple travel modes,''
  \emph{Applied Sciences}, vol.~8, no.~11, p. 2205, 2018.

\bibitem{chen2017socially}
Y.~F. Chen, M.~Everett, M.~Liu, and J.~P. How, ``Socially aware motion planning
  with deep reinforcement learning,'' in \emph{2017 IEEE/RSJ International
  Conference on Intelligent Robots and Systems (IROS)}.\hskip 1em plus 0.5em
  minus 0.4em\relax IEEE, 2017, pp. 1343--1350.

\bibitem{chen2017decentralized}
Y.~F. Chen, M.~Liu, M.~Everett, and J.~P. How, ``Decentralized
  non-communicating multiagent collision avoidance with deep reinforcement
  learning,'' in \emph{2017 IEEE international conference on robotics and
  automation (ICRA)}.\hskip 1em plus 0.5em minus 0.4em\relax IEEE, 2017, pp.
  285--292.

\bibitem{everett2018motion}
M.~Everett, Y.~F. Chen, and J.~P. How, ``Motion planning among dynamic,
  decision-making agents with deep reinforcement learning,'' in \emph{2018
  IEEE/RSJ International Conference on Intelligent Robots and Systems
  (IROS)}.\hskip 1em plus 0.5em minus 0.4em\relax IEEE, 2018, pp. 3052--3059.

\bibitem{chen2019crowd}
C.~Chen, Y.~Liu, S.~Kreiss, and A.~Alahi, ``Crowd-robot interaction:
  Crowd-aware robot navigation with attention-based deep reinforcement
  learning,'' in \emph{2019 international conference on robotics and automation
  (ICRA)}.\hskip 1em plus 0.5em minus 0.4em\relax IEEE, 2019, pp. 6015--6022.

\bibitem{mun2022occlusion}
Y.-J. Mun, M.~Itkina, S.~Liu, and K.~Driggs-Campbell, ``Occlusion-aware crowd
  navigation using people as sensors,'' in \emph{2023 IEEE International
  Conference on Robotics and Automation (ICRA)}.\hskip 1em plus 0.5em minus
  0.4em\relax IEEE, 2023, pp. 12\,031--12\,037.

\bibitem{biswas2013localization}
J.~Biswas and M.~M. Veloso, ``Localization and navigation of the cobots over
  long-term deployments,'' \emph{The International Journal of Robotics
  Research}, vol.~32, no.~14, pp. 1679--1694, 2013.

\bibitem{hirose2022exaug}
N.~Hirose, D.~Shah, A.~Sridhar, and S.~Levine, ``Exaug: Robot-conditioned
  navigation policies via geometric experience augmentation,'' in \emph{2023
  IEEE International Conference on Robotics and Automation (ICRA)}.\hskip 1em
  plus 0.5em minus 0.4em\relax IEEE, 2023, pp. 4077--4084.

\bibitem{savinov2018semi}
N.~Savinov, A.~Dosovitskiy, and V.~Koltun, ``Semi-parametric topological memory
  for navigation,'' in \emph{International Conference on Learning
  Representations}, 2018.

\bibitem{meng2020scaling}
X.~Meng, N.~Ratliff, Y.~Xiang, and D.~Fox, ``Scaling local control to
  large-scale topological navigation,'' in \emph{2020 IEEE International
  Conference on Robotics and Automation (ICRA)}.\hskip 1em plus 0.5em minus
  0.4em\relax IEEE, 2020, pp. 672--678.

\bibitem{shah2022viking}
D.~Shah and S.~Levine, ``Viking: Vision-based kilometer-scale navigation with
  geographic hints,'' \emph{Robotics: Science and Systems}, 2022.

\bibitem{kim2022topological}
N.~Kim, O.~Kwon, H.~Yoo, Y.~Choi, J.~Park, and S.~Oh, ``Topological semantic
  graph memory for image-goal navigation,'' in \emph{Conference on Robot
  Learning}.\hskip 1em plus 0.5em minus 0.4em\relax PMLR, 2023, pp. 393--402.

\bibitem{shah2022gnm}
D.~Shah, A.~Sridhar, A.~Bhorkar, N.~Hirose, and S.~Levine, ``Gnm: A general
  navigation model to drive any robot,'' in \emph{2023 IEEE International
  Conference on Robotics and Automation (ICRA)}.\hskip 1em plus 0.5em minus
  0.4em\relax IEEE, 2023, pp. 7226--7233.

\bibitem{redmon2016you}
J.~Redmon, S.~Divvala, R.~Girshick, and A.~Farhadi, ``You only look once:
  Unified, real-time object detection,'' in \emph{Proceedings of the IEEE
  conference on computer vision and pattern recognition}, 2016, pp. 779--788.

\bibitem{yolo_v5}
``Object detection by yolov5,'' \url{https://github.com/ultralytics/yolov5}.

\bibitem{wojke2017simple}
N.~Wojke, A.~Bewley, and D.~Paulus, ``Simple online and realtime tracking with
  a deep association metric,'' in \emph{2017 IEEE international conference on
  image processing (ICIP)}.\hskip 1em plus 0.5em minus 0.4em\relax IEEE, 2017,
  pp. 3645--3649.

\bibitem{yolo_deepsort}
``Pedestrian detection and tracking by yolov5 and deepsort,''
  \url{https://github.com/HowieMa/DeepSORT_YOLOv5_Pytorch}.

\bibitem{niwa2022spatio}
T.~Niwa, S.~Taguchi, and N.~Hirose, ``Spatio-temporal graph localization
  networks for image-based navigation,'' in \emph{2022 IEEE/RSJ International
  Conference on Intelligent Robots and Systems (IROS)}.\hskip 1em plus 0.5em
  minus 0.4em\relax IEEE, 2022, pp. 3279--3286.

\bibitem{ros_armarker}
``Ros wrapper for alvar, an open source ar tag tracking library,''
  \url{http://wiki.ros.org/ar_track_alvar}.

\bibitem{chebotar2021actionable}
Y.~Chebotar, K.~Hausman, Y.~Lu, T.~Xiao, D.~Kalashnikov, J.~Varley, A.~Irpan,
  B.~Eysenbach, R.~Julian, C.~Finn, \emph{et~al.}, ``Actionable models:
  Unsupervised offline reinforcement learning of robotic skills,'' \emph{arXiv
  preprint arXiv:2104.07749}, 2021.

\bibitem{ros_roomba}
``{Roomba Drivers},'' \url{https://github.com/AutonomyLab/create_robot}.

\bibitem{Tsoi_2021_Sean_EP}
N.~Tsoi, M.~Hussein, O.~Fugikawa, J.~Zhao, and M.~V{\'a}zquez, ``An approach to
  deploy interactive robotic simulators on the web for hri experiments: Results
  in social robot navigation,'' in \emph{2021 IEEE/RSJ International Conference
  on Intelligent Robots and Systems (IROS)}.\hskip 1em plus 0.5em minus
  0.4em\relax IEEE, 2021, pp. 7528--7535.

\bibitem{anderson2018evaluation}
P.~Anderson, A.~Chang, D.~S. Chaplot, A.~Dosovitskiy, S.~Gupta, V.~Koltun,
  J.~Kosecka, J.~Malik, R.~Mottaghi, M.~Savva, \emph{et~al.}, ``On evaluation
  of embodied navigation agents,'' \emph{arXiv preprint arXiv:1807.06757},
  2018.

\bibitem{francis2023principles}
A.~Francis, C.~P{\'e}rez-d'Arpino, C.~Li, F.~Xia, A.~Alahi, R.~Alami, A.~Bera,
  A.~Biswas, J.~Biswas, R.~Chandra, \emph{et~al.}, ``Principles and guidelines
  for evaluating social robot navigation algorithms,'' \emph{arXiv preprint
  arXiv:2306.16740}, 2023.

\bibitem{wang2022metrics}
J.~Wang, W.~P. Chan, P.~Carreno-Medrano, A.~Cosgun, and E.~Croft, ``Metrics for
  evaluating social conformity of crowd navigation algorithms,'' in \emph{2022
  IEEE International Conference on Advanced Robotics and Its Social Impacts
  (ARSO)}.\hskip 1em plus 0.5em minus 0.4em\relax IEEE, 2022, pp. 1--6.

\bibitem{giesbrecht2009vision}
J.~L. Giesbrecht, H.~K. Goi, T.~D. Barfoot, and B.~A. Francis, ``A vision-based
  robotic follower vehicle,'' in \emph{Unmanned Systems Technology XI}, vol.
  7332.\hskip 1em plus 0.5em minus 0.4em\relax SPIE, 2009, pp. 451--462.

\end{thebibliography}
\endgroup
\vfill
\newpage
\section*{Appendix}
\subsection{Localization with Long-term Anchors} \label{sec:app_loc}
%
%Localization module often uses the image similarity or estimated relative poses between current image and node images in the topological map. However, we often have visually similar but different location inside of the buildings. If these scenes are included in the topological map, the localization module often fails and breaks navigation. Although we can practically search for the next limited nodes in the path not to search the visually similar scenes, that causes to accumulate the estimation error and occurs navigation failure. 
%Especially, we often have failures related to localization error in the case without sufficient continuous learning.

In order to avoid navigation failure by the localization errors, we placed some AR tags along the topological graph to assist localization. Our idea is simply overriding the estimated node number by the node number associated with the AR tags. 
When collecting the topological graph, we also save the list of $\{n^{ar}_i, p^{ar}_i, n^{node}_i \}_{i=1 \ldots N_{ar}}$. Here $n^{ar}_i$ and $p^{ar}_i$ are the detected AR tag number and its pose on the robot coordinate, respectively~\cite{ros_armarker}. $n^{node}_i$ is the node number on the topological graph, which detects the AR tag of $n^{ar}_i$.

We basically override the estimated node number by $n^{node}_j$ when detecting AR tag of $n^{ar}_i$ in the data collection. If the multiple node images detect the same AR tag in the topological graph, we use the closest one to assist moving forward.
%move forward. It can avoid to go back and be away from the goal position in navigation. 
However, the mobile robot may pass over the node location linked to the AR tag and still detect its AR tag. 
%In such a case, the subgoal image behind the current robot pose will be given. 
Such a case causes unnatural movement like stopping abruptly because the subgoal image will be behind the current robot pose. To avoid unnatural behavior, the robot compares the estimated pose of AR tag with $p^{ar}_i$ to detect whether the its is passing by the tag. If the robot is passing by, it is overwritten with the next node number $n^{ar}_i$ + 1.
\begin{figure}[h]
  \begin{center}
      \includegraphics[width=0.99\hsize]{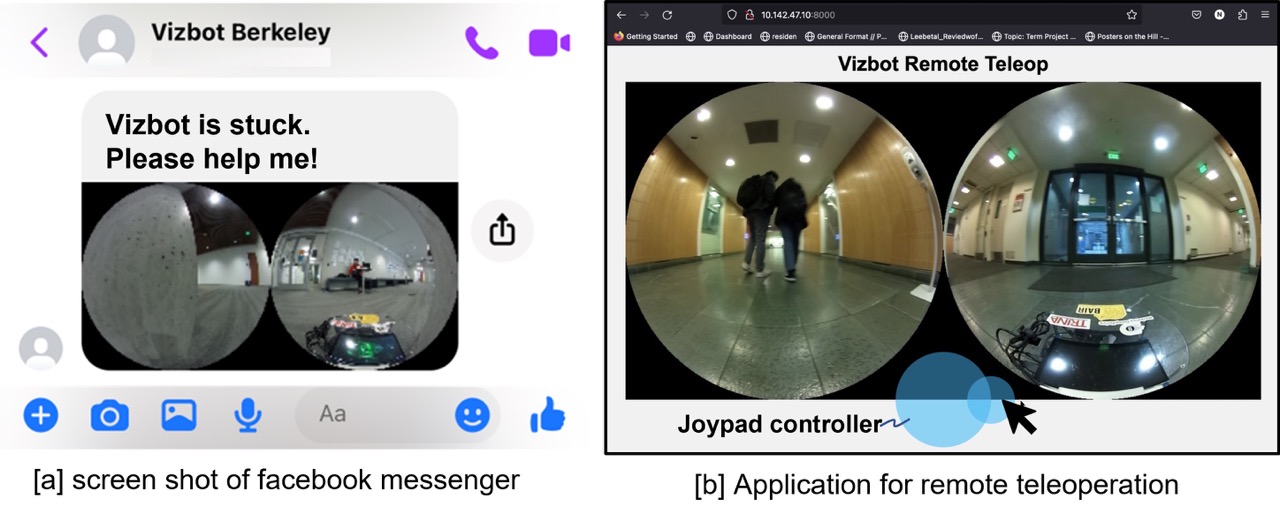}
  \end{center}
      \vspace*{0mm}
	\caption{\small {\bf Help-and-rescue module.} The robot messages the operator for assistance when stuck (left), and can be rescued remotely over the internet using our web interface (right).}
  \label{f:rescue}
  \vspace*{0.0em}
\end{figure}
\subsection{Help-and-rescue module} \label{sec:help_rescue}
In the help-and-rescue module, we implement a pipeline for autonomous recovery from collisions and seeking remote help in case of irrecoverable collisions for the challenging obstacles (e.g., that may be shorter than the camera height, made of glass etc.). When a collision is detected by the robot's collision detector sensor (e.g., a mechanical contact sensor), an automatic backup maneuver is executed. This maneuver drives the robot away from the obstacle for a short distance along the normal vector corresponding to the point of contact. Specifically, the robot moves back about 0.5 meter and rotates about 45 degrees at a point. The direction of rotation is determined by the detection of two bumper sensors in right and left. If the left sensor detects a collision, the robot rotates to the right; if the right sensor detects a collision, the robot rotates to the left.

This allows the robot to automatically recover from 70\% of the simple collisions where the robot accidentally runs into challenging obstacles (e.g., that may be shorter than the camera height, made of glass etc.). Complete autonomy, however, may not be possible to achieve. The robot may drive itself into a convex hull of multiple obstacles, leading to repeated collisions, or get it's wheels stuck (e.g., on an air vent) and be unable to rescue itself. Only for these accidental cases, we use a messaging and remote teleoperation interface in Fig.~\ref{f:rescue} to recover and continues the data collection without any physical interventions. 
\subsection{Trajectory Chaining for Continual Learning} \label{sec:app_chain}
\begin{figure}[ht]
  \begin{center}
      \includegraphics[width=0.8\hsize]{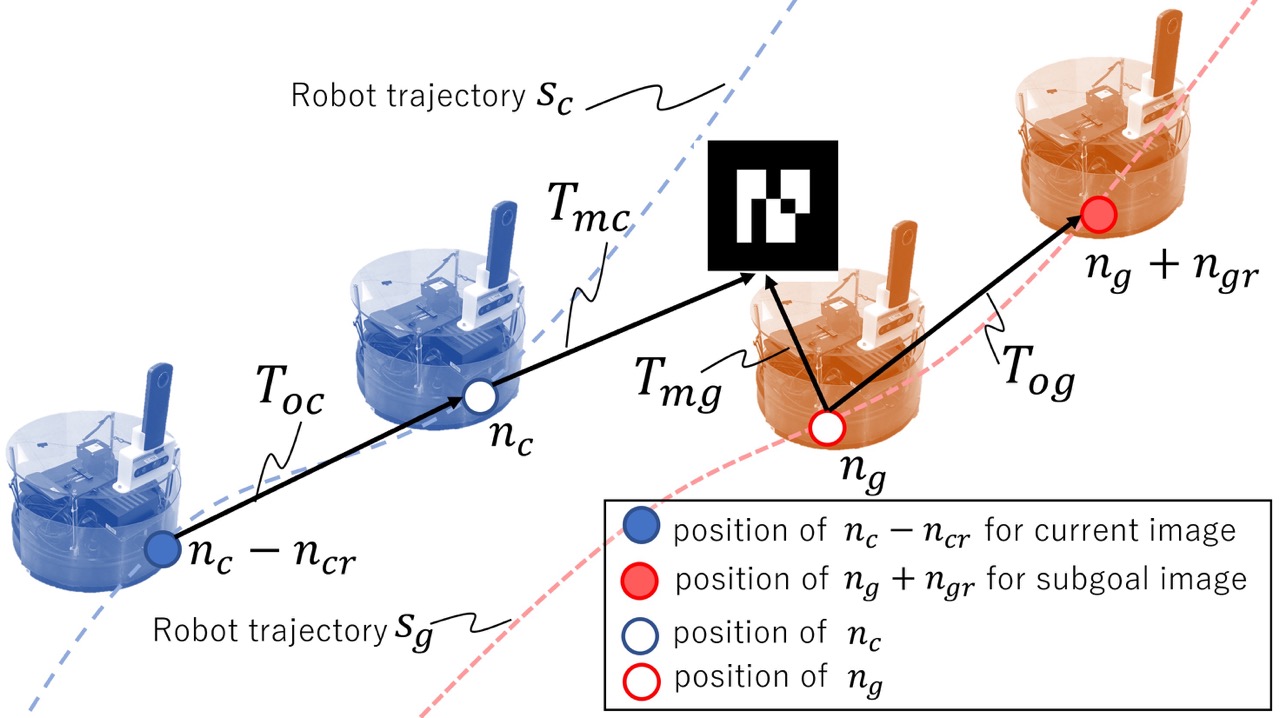}
  \end{center}
      \vspace*{0mm}
	\caption{\small {\bf Relative positions from different sequences using AR tags.} We obtain the relative pose using the odometry of the robot and the detected pose of the AR tag for continuous learning.}
  \label{f:cl_ar}
  \vspace*{0mm}
\end{figure}

To chain the different sequences in training, we need to take $T_{gt}$ between current and subgoal image from different sequences. Fig.~\ref{f:cl_ar} visualizes how to obtain $T_{gt}$ from different sequences. Since we place AR tags along the topological graph to assist the localization module, some frames in our dataset detect AR tag and estimate the relative pose for each AR tag. In Fig.~\ref{f:cl_ar}, $T_{mc}$ and $T_{mg}$ indicate the estimated relative pose against same AR tag from different sequence $s_c$ and $s_g$. Here, $n_c$ and $n_g$ are corresponding node number on $s_c$ and $s_g$. 

To take various pairs of current and subgoal images, we randomly select two step numbers within $N_m = 18$ as $n_{cr}$ and $n_{gr}$ and decide the node number of the current image as $n_c$ - $n_{cr}$ on $s_c$ and the node number of the subgoal image as $n_g$ + $n_{gr}$ on $s_g$, respectively. The sign of $n_{cr}$ and $n_{gr}$ are decided so that the subgoal image position is forward with respect to the current image position. Note that we assume that the dataset can be collected with a positive linear velocity. Since $N_m$ is not large number, we can have accurate relative pose $T_{oc}$ between $n_c - n_{cr}$ and $n_c$, and an accurate relative pose $T_{og}$ between $n_g$ and $n_g + n_{gr}$ from the odometry. As the result, we calculate $T_{gt}$ between $n_c - n_{cr}$ and $n_g + n_{gr}$ as $T_{gt} = T_{oc} \cdot T_{mc} \cdot T_{mg}^{-1} \cdot T_{og}$.
%
%
%\begin{eqnarray}
%    T_{gt} = T_{oc} \cdot T_{mc} \cdot T_{mg}^{-1} \cdot T_{og}.
%    \label{eq:tgt}
%\end{eqnarray}
%

%In addition, we take the images as $I_t$ from $n_c - n_{cr}$, $I_g$ from $n_g + n_{gr}$, and $I_j$ from $n_c - n_{cr} - j$ to calculate our control policy in eq.~(\ref{eq:control_policy}).
%According to the above image selection and $T_{gt}$ in training, we can leverage the data sequence at different date, which naturally includes the environmental changes and can train the robust control policy. 
%
\subsection{Network structures} \label{sec:net_struct}
\begin{figure}[ht]
  \begin{center}
      \includegraphics[width=0.99\hsize]{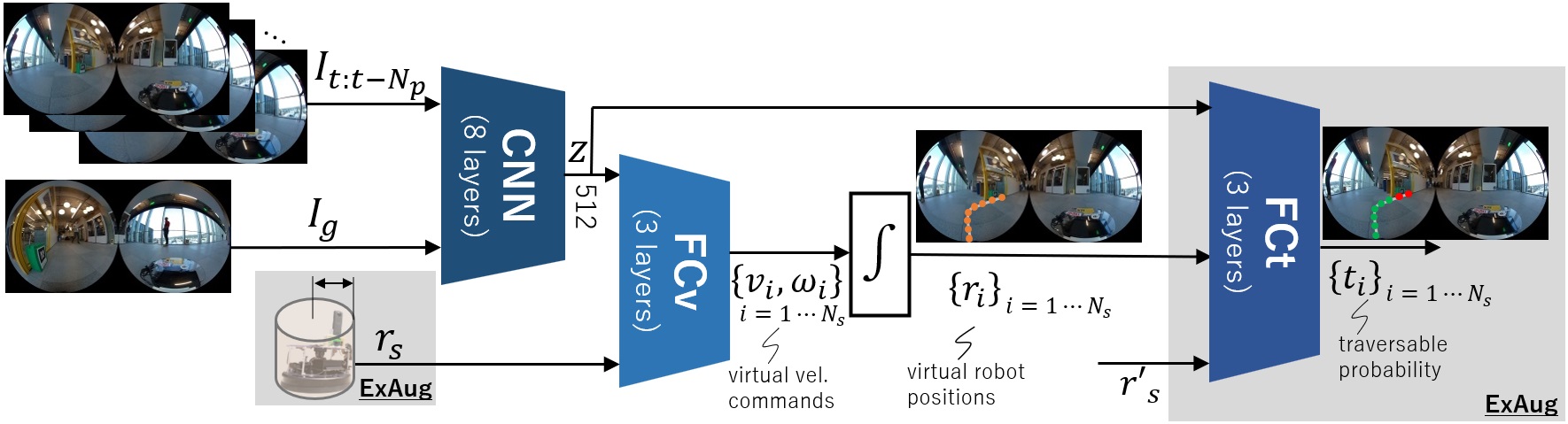}
  \end{center}
      \vspace*{0mm}
	\caption{\small {\bf Network structure of our control policy.}}
  \label{f:network_cp}
  \vspace*{0mm}
\end{figure}

Figure~\ref{f:network_cp} describes the neural network architecture of $\pi_\theta$. An 8-layer CNN is used to extract the image features $z$ from the image history $I_{t:t-N_p}$ and the subgoal image $I_g$, with each layer using BatchNorm and ReLU activations.  
Following our previous work, ExAug, the predicted velocity commands $\{v_i, \omega_i\}_{i=1\ldots N_s}$ from 3 fully-connected layers ``FCv'' are conditioned on the robot size $\{r_s, v_l\}$ and $z$. A scaled tanh activation is given to limit the output velocities as per the specified constraints. We can control the robot by giving $v_1$ and $\omega_1$ as the actual robot velocity command.

In addition to the core part of our control policy, we can implement "FCt" to estimate traversability $\{t_i\}_{i=1\ldots N_s}$, following ExAug. We integrate the velocities to obtain waypoints predictions and feed them to a set of fully-connected layers ``FCt'' along with the observation embedding $z$ and target robot size $r_s'$, followed by a sigmoid function to limit $t_i \in (0, 1)$. Although $r_s = r_s'$ in training, we found the flexibility of an independent $r_s' \neq r_s$ crucial to the collision-avoidance performance of our system in inference. Note that we can remove the gray color part to construct $\pi_\theta$ for the minimum implementation.

\begin{figure}[ht]
  \begin{center}
      \includegraphics[width=0.99\hsize]{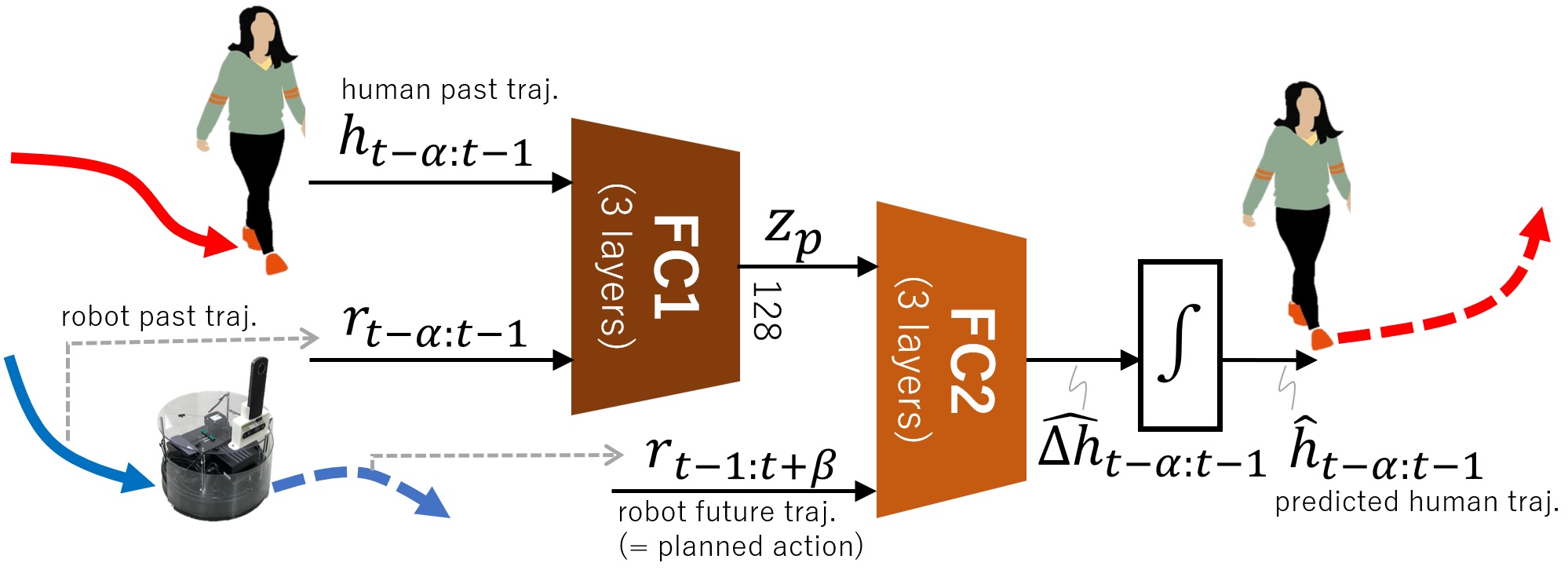}
  \end{center}
      \vspace*{0mm}
	\caption{\small {\bf Network structure of our model to predict pedestrians dynamics.}}
  \label{f:network_pr}
  \vspace*{0mm}
\end{figure}

In our evaluation section, we train the predictive model $f_{\theta_p}$ for the pedestrians dynamics. Fig.~\ref{f:network_pr} is the network structure of $f_{\theta_p}$. At first, we feed the concatenated past human trajectory $h_{t-\alpha:t-1}$ and the past robot trajectory $r_{t-\alpha:t-1}$ into ``FC1'' with the three fully connected layers using BatchNorm and ReLU activations to extract the features $z_p$. Then, we predict the human future trajectory condition on the robot actions (=future trajectories) by giving $z_p$ with $r_{t-1:t+\beta}$. Here, the last layer of ``FC2'' with three fully connected layers has the tanh activation to limit the human velocity within $\pm$ 1.5 m/s.

\subsection{Simulation dataset from social force model}
\label{sec:app_sfm}
To evaluate the effectiveness of SACSoN dataset, we generate the pedestrian trajectories and the robot trajectories from the social force model~\cite{helbing1995social}. In addition, we mix this simulation dataset with the real data from the SACSoN dataset in training to improve the accuracy of the predictive model for the pedestrians' future trajectories. In this appendix, we show the implementation details to generate the simulation dataset.

We set two agents (the robot and the pedestrian) with different initial velocity: 0.8 m/s for the pedestrian and 0.3 m/s for the robot toward the goal position, because the pedestrian is much faster than our robot in our case. Note that the initial velocity decides a nominal velocity for each agent, not a maximum velocity. To simulate these agents in each scenario, we randomly place these agents on their own circle with varying radii, such that the two circles centered at the origin. We decide the robot's and pedestrian's goal position as the opposite side of their respective circles. However, we randomly shorten the goal position for the robot to stop before arriving at the original goal to emulate giving way to the pedestrian. We decide the radius for the robot's circle as 2.0 m and the radius for the pedestrian's circle as 5.3 m. Since the center of these circles is the origin, the robot and the pedestrian often has the interaction around the origin, because the radius for the pedestrian 5.3 m is calculated as $\frac{0.8}{0.3} \times$ 2.0 m. We run the social force model with these hyperparameters for 80 steps and collect 10000 scenarios.

\begin{figure}[ht]
  \begin{center}
      \includegraphics[width=0.99\hsize]{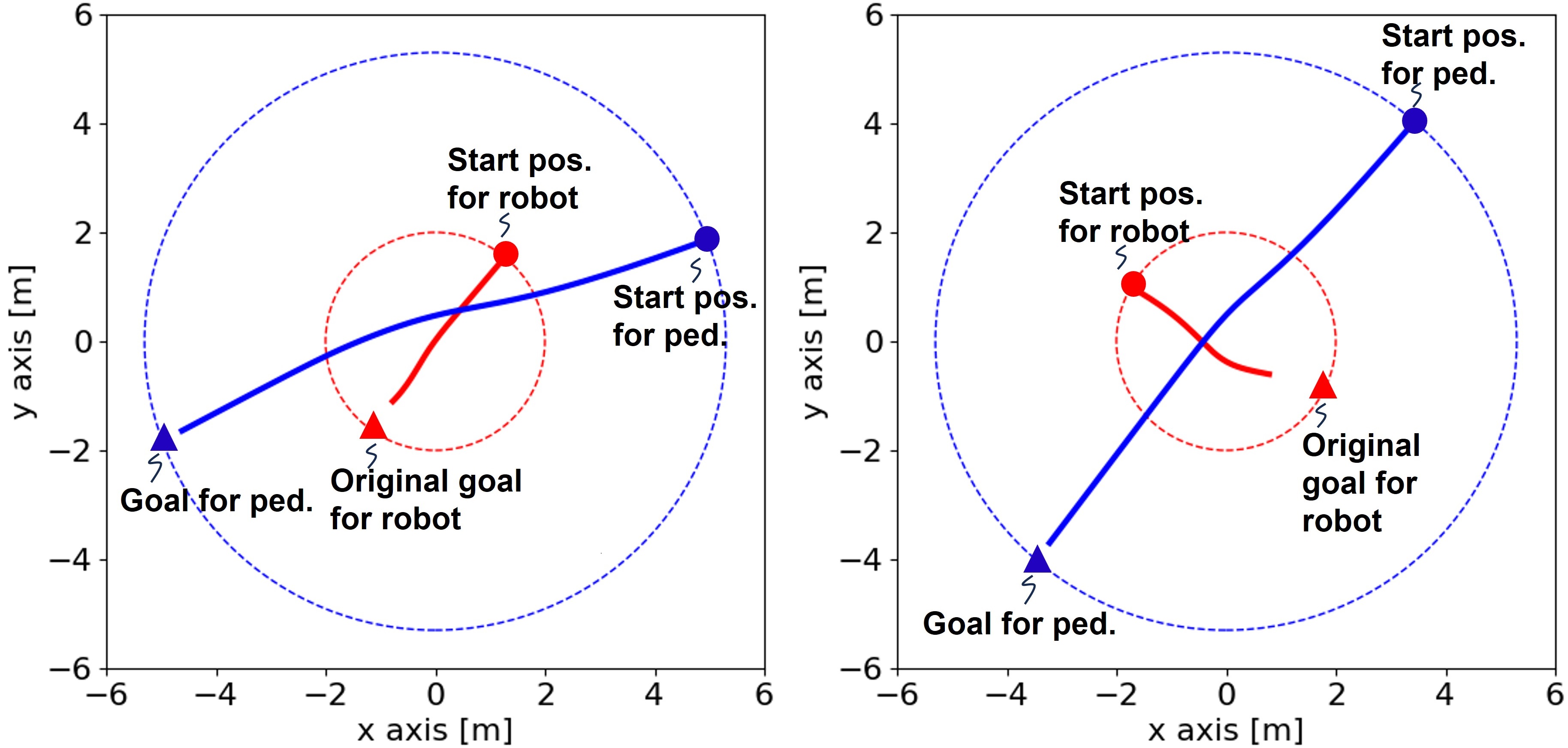}
  \end{center}
      \vspace*{0mm}
	\caption{\small {\bf Examples of simulation dataset from sorcial force model.}}
  \label{f:sf_model}
  \vspace*{0mm}
\end{figure}

In training, we randomly choose the scenario to make the batch. Since our predictive model estimates the pedestrians trajectory on the robot local coordinate, we transform the sampled trajectories before making batch. In Fig.~\ref{f:sf_model}, we show the examples of the simulation dataset from the social force model.
\vfill
\end{document}